\pdfoutput=1
\documentclass[11pt]{article}

\usepackage[final]{acl}

\usepackage{times}
\usepackage{latexsym}
\usepackage{amssymb}
\usepackage{caption}
\usepackage{subcaption}
\usepackage{mathtools}
\usepackage{rotating}
\usepackage{xcolor}
\DeclarePairedDelimiter{\norm}{\lVert}{\rVert} 

\usepackage[T1]{fontenc}
\usepackage[utf8]{inputenc}

\usepackage{microtype}

\usepackage{inconsolata}

\usepackage{graphicx}

\title{Mechanistic Interpretability of Emotion Inference in \\Large Language Models
}

\author{
 \textbf{Ala N. Tak\textsuperscript{1,3,*}},
 \textbf{Amin Banayeeanzade\textsuperscript{2,3,*}},
 \textbf{Anahita Bolourani\textsuperscript{4}},
\\
 \textbf{Mina Kian\textsuperscript{3}},
 \textbf{Robin Jia\textsuperscript{3}},
 \textbf{Jonathan Gratch\textsuperscript{1,3}},
\\
 \textsuperscript{1}Institute for Creative Technologies, University of Southern California (USC),
\\
\textsuperscript{2}
Information Sciences Institute, University of Southern California (USC),
\\
 \textsuperscript{3}Department of Computer Science, University of Southern California (USC),
\\
 \textsuperscript{4}Department of Statistics and Data Science, University of California, Los Angeles (UCLA)
\\
 \small{
   \textbf{Correspondence:} \href{mailto:antak@ict.usc.edu}{antak@ict.usc.edu}; \href{mailto:banayeea@usc.edu}{banayeea@usc.edu}, \textsuperscript{*}Equal contribution
 }
}

\begin{document}
\maketitle
\begin{abstract}

% We used Mechanistic Interpretability to reveal localized emotion representations in LLMs, validate their psychological alignment, and demonstrate causal control over emotion inference for better interpretability.

Large language models (LLMs) show promising capabilities in predicting human emotions from text. However, the mechanisms through which these models process emotional stimuli remain largely unexplored. Our study addresses this gap by investigating how autoregressive LLMs infer emotions, showing that emotion representations are functionally localized to specific regions in the model. Our evaluation includes diverse model families and sizes, and is supported by robustness checks. We then show that the identified representations are psychologically plausible by drawing on cognitive appraisal theory—a well-established psychological framework positing that emotions emerge from evaluations (appraisals) of environmental stimuli. By causally intervening on construed appraisal concepts, we steer the generation and show that the outputs align with theoretical and intuitive expectations. This work highlights a novel way to causally intervene and control emotion inference, potentially benefiting safety and alignment in sensitive affective domains. \textit{Code at: \href{https://github.com/aminbana/emo-llm}{\texttt{GitHub repo.}}}

% keywords: Mechanistic Interpretability, emotion detection and analysis, cognitive modeling

\end{abstract}

\section{Introduction}

\begin{figure}[!t]
\centering
  \includegraphics[width=.95\columnwidth]{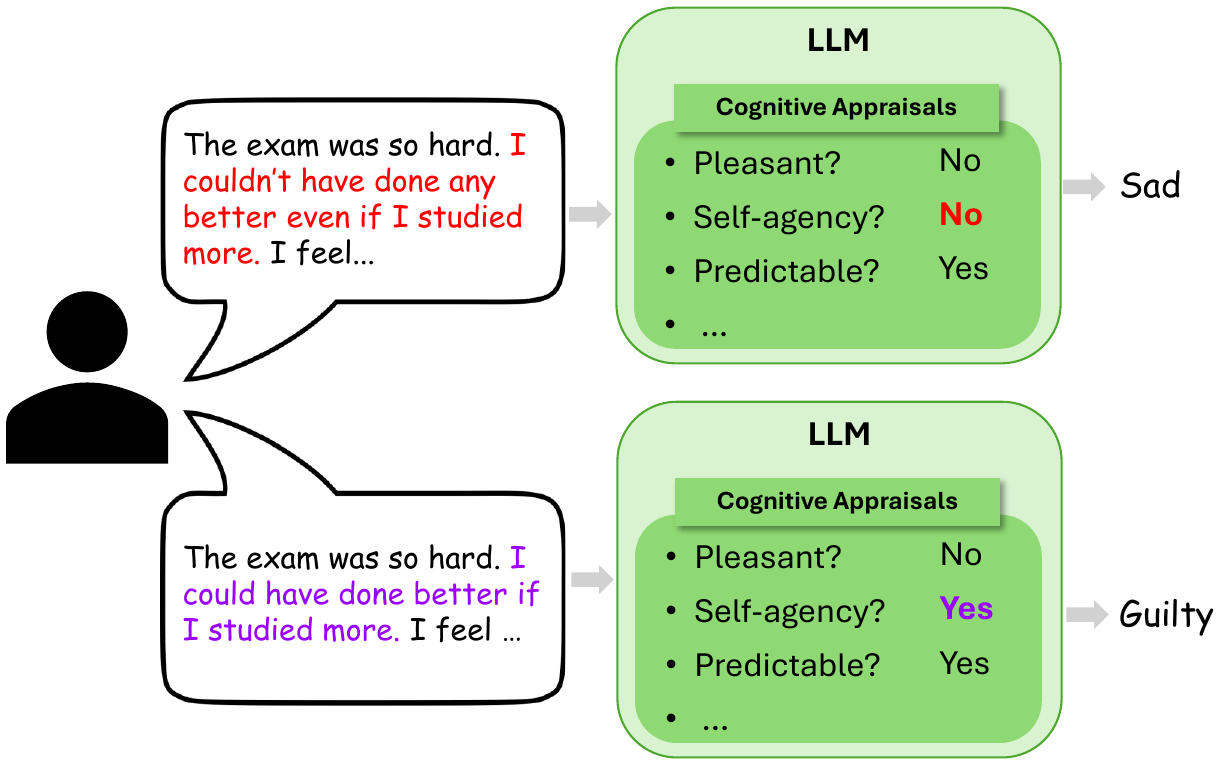}
  \caption{
  Emotion inference through latent appraisal-like mechanisms in LLMs. Given the description of a situation, the model leverages internal \textit{appraisal} structures to recognize the emotion inferred from the context. For example, different perceptions of \textit{self-agency} can distinguish between \textit{guilt} and \textit{sadness}.
  }
  \vspace{-5mm}
  \label{fig:appraisal}
\end{figure}

Large Language Models (LLMs) demonstrate remarkable capabilities in emotion recognition and reasoning tasks, occasionally surpassing human performance \cite{elyoseph2023chatgpt, tak2024gpt4emulatesaveragehumanemotional}. Prior research primarily engages with LLMs as black boxes, utilizing zero-shot inference or in-context learning to gauge their performance on tasks such as emotion classification \cite{yongsatianchot2023investigating, broekens2023fine}, emotional decision-making and situational appraisal \cite{10388119}, emotional intelligence \cite{wang2023emotional}, emotional dialogue understanding \citep{zhao2023chatgpt}, and generation of emotional text \cite{gagne2023inner}.
However, there remains a limited understanding of \emph{how} LLMs internally represent and process emotional information. Given LLMs' increasingly significant societal impact—spanning domains such as mental health \cite{2023_cognitive} and legal decision-making \cite{lai2024large}—investigating these internal mechanisms is crucial.

Cognitive neuroscience uses functional localization approaches to identify specific brain regions responsible for particular functions and manipulate them by up/down-regulating neural activations in those regions. Akin to the shift from behaviorism to cognitive neuroscience in psychology—i.e. from treating the mind as a black box to studying brain-based cognitive processes—Mechanistic Interpretability (MI) allows for moving from black-box techniques \cite{casper2024black}, to a focus on the internal mechanics of LLMs \cite{bereska2024mechanisticinterpretabilityaisafety}. MI can offer a fundamental understanding of information processing and representation in LLMs, yielding insights into their inner-workings and offering new ways to control their reasoning \cite{li2021implicit, rai2024practical, feng2024monitoringlatentworldstates}. Building on this line of research and by drawing inspiration from emotion theory in psychology, our work elucidates the inner workings of emotion processing in LLMs. 

We start by training linear classifiers on top of hidden representations to probe for regions where the strongest emotion-related activations occur. We provide evidence for functional localization of emotion processing and show that emotion-relevant operations are concentrated in specific layers, a consistent behavior across various model families and scales. We complement these findings by applying causal interventions, namely patching activations in the computation graphs, to identify essential components in neural representations \cite{conmy2023towards}. 
%specifically, we inspect the role of architectural components such as Multi-Head Self-Attention (MHSA) and Feed-Forward Network (FFN) \cite{geva2022transformer, 2024circuitreuse}. 
As a result, we show that Multi-Head Self-Attention (MHSA) units in the mid-layers are predominantly responsible for shaping the LLM decision. To further corroborate this, we visualize attention patterns, revealing that MHSA units consistently attend to emotionally loaded tokens. Our findings are robust and not influenced by variations in prompt wording or formatting.

Additionally, we use the \textit{appraisal theory} from psychology to shed light on the structure of LLMs' internal representations. According to appraisal theory \cite{frijda1989relations, scherer1984nature, smith1985patterns}, people reason about emotional situations by forming \textit{appraisal} judgments (see Figure~\ref{fig:appraisal} as an example). We analyze the structure of emotion representations in LLMs by conducting inference-time probing on appraisal concepts to show that representations are psychologically plausible. Moreover, by modulating latent appraisal concepts to promote/demote a particular appraisal dimension (e.g., promoting \textit{self-agency}), we show that the resulting changes in the output emotion align with theoretical expectations (from \textit{guilt} to \textit{sadness}) \cite{laguna2024beyond, wu2024replymakelovetal, li2024inferencetimeinterventionelicitingtruthful}. 

Overall, our work extends existing MI methodologies by applying them to more ecologically valid, unstructured examples, moving beyond the common practice of analyzing simplified sentence structures \cite{wang2022interpretability, hanna2023doesgpt2}. Furthermore, this work serves as an early step to bridge MI techniques with applications in psychological and cognitive domains, offering insights into the inner workings of LLMs in complex, socially relevant contexts.

\section{Related Work}

\textbf{Appraisal Theory.}
Appraisal theory is a model that explains how peoples' emotions are a result of their evaluations of a situation \cite{lazarus1991emotion}. It provides a comprehensive framework for understanding the precursors of emotions \cite{smith2011role}. Neuroscience studies build on this framework by manipulating cognitive appraisals and examining associated brain activity, linking specific brain regions to appraisal processes \cite{leitao2020computational, kragel2024can, brosch2013comment}. These methods can be extended to evaluate how LLMs understand emotions, and identify the mechanisms responsible for those evaluations. 

% \subsection{LLMs' Emotion Capabilities}
% In their review, \citet{Yongsatianchot} investigate the emotion capabilities of LLMs. They identify two ongoing areas of research exploring the use of LLMs: emotion recognition and emotion generation. Within the domain of emotion recognition, many studies have approached the evaluation of LLMs through the perspective of appraisal \cite{yongsatianchot2023investigating,zhan2023evaluating,yongsatianchot2023s}, emotional intelligence \cite{wang2023emotional,paech2023eq,croissant2024appraisal,zhao2024both}, or emotional awareness \cite{elyoseph2023chatgpt}  theories. LLMs have also been evaluated in emotion generation tasks \cite{broekens2023fine,coda2023inducing,croissant2024appraisal,li2023large, vu2024psychadapter}. In addition to these ongoing areas of evaluation, \citet{Yongsatianchot} recommend that future research investigate \textit{how} these models are able to approach emotion-related tasks.

% % Croissant et al. (2023) chain of emotion prompting 

\textbf{Cognitive-Neuroscientific Alignment.} There is a growing body of evidence demonstrating that neural network activations can reflect cognitive constructs traditionally studied in psychology and neuroscience. For example, transformer self-attention mechanisms are shown to correlate with human eye-tracking data during reading, suggesting that LLMs may learn attention patterns that reflect human cognitive attention \cite{bensemannetal2022eye, eberleetal2022transformer}. LLMs also exhibit strong representational alignment with neural activity in the human language systems, indicating convergence between learned representations and brain-like processing \cite{aw2024instructiontuning}, and word embeddings show alignment with fMRI activity during word association tasks \cite{osti_10560521}. Recent work even uses LLM-derived representations to predict neural activation patterns tied to high-level semantic categories like faces or places, offering scalable proxies for annotation in cognitive experiments \cite{liu2025talkingbrainusinglarge}. Other studies localize LLM units that align with cortical language areas \cite{alkhamissi2024llm} and reveal functional specialization patterns predictive of regional brain activity \cite{kumar2024shared}.

These findings support the plausibility of mapping psychological or cognitive theories—such as appraisal theory—onto the internal structure of LLMs. Our work builds on this foundation by examining how appraisal mechanisms are internally encoded, bridging prior behavioral studies of appraisal-emotion mappings in LLMs \cite{tak2024gpt4emulatesaveragehumanemotional, 10388119,  broekens2023fine, yongsatianchot2023investigating} with MI.

\textbf{Mechanistic Interpretability.}
 % Mechanistic interpretability (MI) aims to analyze the internal structures of models and understand how those structures are leveraged to perform various tasks \cite{rai2024practical}. %With the advent of increasingly more powerful LLMs, an emerging field in NLP focuses on understanding how these models perform various tasks.
 %As a result, various MI techniques are applied to assess transformer-based language models \cite{rai2024practical}.
 %A variety of MI methods are proposed to investigate the representations in LLMs including probing \cite{prob2019, belinkov2022probing},
 %circuit discovery \cite{nanda2023progress, wang2022interpretability, conmy2023towards, meng2022locating, prakash2024finetuningenhancesexistingmechanisms}
 %patching \cite{heimersheim2024use}, and generation steering \cite{templeton2024scaling}. 
Probing is an MI technique that uses a simple model, called a “probe”, to assess the internal representations across various layers in a model. As explained by \citet{belinkov2018internal}, the groundwork for what we now refer to as probing relates back to earlier work evaluating trained classifiers on static word embeddings to predict linguistic features \cite{kohn2015s,gupta2015distributional}, and classified hidden states of neural models \cite{ettinger2016probing,kadar2017representation,shi2016does,adi2016fine,hupkes2018visualisation,belinkov2022probing, giulianelli2018under}. Probing is used across a variety of tasks \cite{hewitt2019designing, tenney2019bert,tenney2019you,peters2018dissecting,clark2019does,belinkov2018internal,conneau2018you}.

Activation patching \cite{heimersheim2024use}, is a causal intervention used to identify if certain activations are important to the downstream task \cite{vig2020investigating}. By using patching, \citet{meng2022locating} are able to localize where models store factual information.
%The process of activation patching works as follows: first, develop two similar prompts, a source prompt and a destination prompt, then run the model with the source prompt. Next, run the model again with the destination prompt, but overwrite the internal activations with ones saved from the original run with the source prompt at key points in the model. Finally, observe the changes to the model output. This process is repeated for all activations of interest. 
Patchscope, a method that extends on activation patching, is used to translate LLM representations into natural language \cite{patchscopes}. 
%Testing with Concept Activation Vectors (TCAV) leverages high-level interpretations of a models internal state in terms of human-friendly concept (Kim et al. 2024). While originally developed for image processing, it has also been used as an approach for evaluations LLMs as well ()

% and show that different internal “circuits” are activated under zero-shot and few-shot conditions.

Yet another MI technique is generation steering. This method entails manipulating a model's activations to control the outputs \cite{rai2024practical, todd2024function}. \citet{geva2022transformer} highlight the role of Feed-Forward Network (FFN) units in promoting concepts. To steer generation, they apply sub-updates promoting safety and are able to reduce the model's toxicity. 
%%\citet{templeton2024scaling} find that feature steering can be used to control the models' output. By "clamping" on the features for the Golden Gate Bridge, the model began to self-identify as the popular landmark in it's generated responses. They were also able to steer the model's stated goals, biases, and  
\citet{templeton2024scaling} find that "clamping" on features can be used to control the models' output, steering the model's stated goals and biases for both desirable and undesirable outputs. 
\citet{nanda2023emergent} demonstrate that sequence models can have linear internal representations and that these representations can be used to manipulate the model's behavior. This method closely resembles those of \citet{turner2308activation} and \citet{lieberum2023does}.

%inference time intervention 

\begin{figure*}[ht!]
  \includegraphics[width=\linewidth]{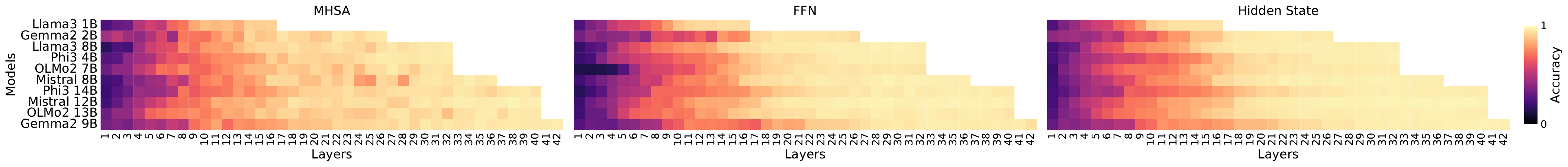}
  \caption {Layer-wise accuracies of emotion probe experiments across different models (each row) with varying depths at (\textbf{Left}) MHSA, (\textbf{Mid}) FFN, and (\textbf{Right}) hidden states. The results suggest an increasing signal with clear consolidation in the mid layers across various model families and sizes, which indicates that models predominantly make emotion-related decisions by the mid layers with minimal improvement in higher layers. %FFN outputs closely follow the hidden states, while attention accuracies exhibit some fluctuations even in the upper layers.
}
  \label{fig:e_probe_heat}
\end{figure*}
\section{Experimental Setup}

\textbf{Dataset and Prompt Design.}
We employ the crowd-enVENT dataset developed by \citet{envent}, which comprises 6,800 emotional vignettes annotated with self-reported emotions among a list of 13 options and 23 self-rated appraisal variables, reflecting nuanced stimuli evaluations, including: \textit{pleasantness/unpleasantness}, \textit{self-agency/other-agency}, \textit{predictability/suddenness}. Appendix~\ref{app:dataset} presents more details on the dataset, including a detailed list of appraisal variables, along with the scales used for measurement. 

To evaluate the model's ability to infer emotions from textual contexts, we design prompts that guide the model to predict the appropriate emotion as the next immediate output token, framing the task as a causal language modeling problem. Subsequently, we consider a classification problem and evaluate the model by inspecting the logits confined to the set of targeted emotion labels. The primary prompt template used in this study is shown in Figure~\ref{fig:aggregate_attention_viz}. 

Emotion attribution is inherently subjective, making it challenging to define a single ground truth label for each input, particularly given our fine-grained list of emotions. Thus, we focus on the correctly classified examples when inspecting each language model. In other words, we only analyze the samples for which the LLM and the human annotator agreed on the same label, totaling at least 2,700 samples among different language models (see Appendix~\ref{app:task} for more details). This ensures that we assess tasks where the model performs reliably to understand the underlying mechanisms. %However, we acknowledge that some emotion classes have a limited number of examples after filtering, which may present constraints in our experiments. 

% , including, \textit{anger, boredom, disgust, fear, guilt, joy, neutral, pride, relief, sadness, shame, surprise, trust}. 
% . To identify the most significant factors for analysis, we employ Factor Analysis, reducing these variables to primary appraisal dimensions 

\textbf{Model Architecture.} To account for the impact of model scale and architectural variations, we evaluate a diverse set of model families and sizes, including Llama~3.2~1B Instruct and Llama~3.1~8B~Instruct \cite{llama3}, Gemma~2~2B~Instruct and Gemma~2~9B~Instruct \cite{gemma2}, OLMo~2~7B~Instruct and OLMo~2~13B~Instruct \cite{olmo2}, Phi~3.5~mini~Instruct and Phi~3~medium-Instruct \cite{abdin2024phi}, and Ministral~8B~Instruct and Mistral~12B~Nemo~Instruct \cite{mistral2024nemo} (see Appendix~\ref{app:arch} for more details). Some detailed analyses, robustness tests, and appraisal concept interventions are exclusively conducted on Llama~3.2~1B to manage computational resources effectively.

\section{Notations and Preliminaries} \label{sec:notations}

% To locate the most salient emotion-related information in the language model, we specify several potential activation extraction locations within each transformer layer. 
Prior research suggests that both MHSA and FFN units drive the generation in specific downstream tasks such as indirect object identification or concept promotion \cite{2024circuitreuse, geva2022transformer}. By examining activations immediately after these units, we aim to evaluate their respective contributions to emotion processing within each transformer layer. More formally, let \( \mathbf{h}^{(l)}_t \in \mathbb{R}^d \) denote the hidden state vector at layer \( l \) and token index \( t \in \{1, \cdots, T\} \), where \( d \) is the dimensionality of the model's hidden representations and $T$ is the input sequence length. Then,

\[
\mathbf{a}^{(l)}_t = \mathbf{MHSA}(\mathbf{h}^{(l-1)}_{1:t}),
\] \vspace{-1mm}
\[
\mathbf{m}^{(l)}_t = \mathbf{FFN}(\mathbf{h}^{(l)}_t + \mathbf{a}^{(l)}_t),
\]\vspace{-1mm}
\[
\mathbf{h}^{(l)}_t = \mathbf{h}^{(l-1)}_t + \mathbf{a}^{(l)}_t + \mathbf{m}^{(l)}_t,
\]

where \( \mathbf{a}^{(l)}_t \in \mathbb{R}^d \) and \( \mathbf{m}^{(l)}_t \in \mathbb{R}^d \) are MHSA and FFN outputs at layer \( l \) for token \( t \). \( \mathbf{h}^{(l-1)}_{1:t} \) are the previous layer's hidden states for tokens 1 to \( t \). 

Throughout this paper, we focus on activations $\mathbf{x}_t^{(l)}$ selected from one of the three candidates in $\{ \mathbf{a}_t^{(l)}, \mathbf{m}_t^{(l)}, \mathbf{h}_t^{(l)} \}$ and study their properties at different layers and token positions. While activations can be extracted from any token index, we anticipate the strongest emotion signals to be present at the last token, as it directly influences the model's output prediction in our task setup. Therefore, when clear from context, we omit the subscript $T$ while studying the last token. Also, we drop the superscript $(l)$ when generally discussing any activation across different layers.

\begin{figure*}[t!]
  \includegraphics[width=\linewidth]{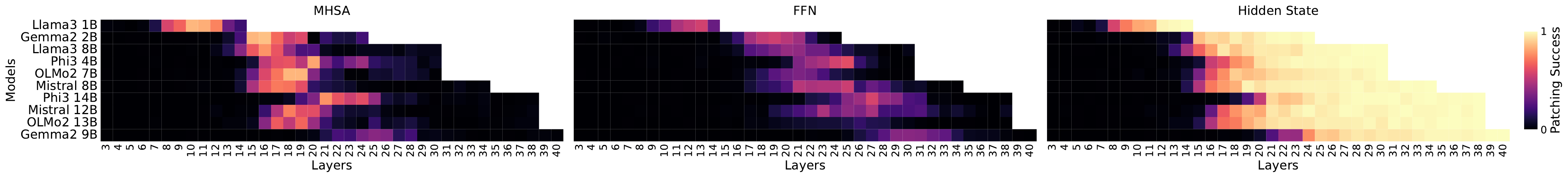}
  \caption {Results of activation patching experiments where we measure the success rate of transferring the output emotion by patching an activation from a source sample to a target sample. The patching is performed at (\textbf{Left}) MHSA, (\textbf{Mid}) FFN, and (\textbf{Right}) hidden states, respectively. The MHSA and FFN heatmaps demonstrate a clear localization with successful transfer peaks happening in the mid-layers consistently across various model families and scales. This observation aligns with the consolidation points observed in the probe heatmap (Figure \ref{fig:e_probe_heat}) and indicates how activation patching can identify critical regions for emotion prediction.
}
  \label{fig:e_patch_heat}
  \vspace{-2mm}
\end{figure*}
\section{Probing for Emotion Signals} \label{sec:emotion_probe}

Building on the linear representation hypothesis \cite{mikolov-etal-2013-linguistic, elhage2022toymodelssuperposition, parklinearhype}, we perform probing experiments to assess the presence and strength of emotion-related signals at different activations within the model. Specifically, we train linear classifiers \cite{hewitt2019designing} to predict the corresponding emotions. We formalize the linear classifiers as follows:

    \[
    \hat{\mathbf{y}} = \mathbf{W}^\top \mathbf{x} + \mathbf{b},
    \]

where \( \mathbf{x} \in \mathbb{R}^d \) denotes the activation vectors at one of the locations specified in the previous section. \( \mathbf{W} \in \mathbb{R}^{d \times C} \) is the weight matrix for emotion classification, \( \mathbf{b} \in \mathbb{R}^C \) is the bias vector, \( C \) represents the number of emotion classes, and \( \hat{\mathbf{y}} \in \mathbb{R}^C \) denotes the predicted logits for each emotion class.

We perform probing separately over different activation locations and layers across the model for the last token index. The probing results in Figure~\ref{fig:e_probe_heat}, measured as the accuracy on a held-out test set, indicate that the models begin consolidating emotional information in the hidden states $\mathbf{h}^{(l)}$ neither too early nor too late, but predominantly around the mid-layers across all models. For example, in the first row corresponding to Llama~3.2~1B in Figure~\ref{fig:e_probe_heat}, the emotional signal peaks by layer \( {l = 10} \) out of a total of $16$ layers. Beyond layer $10$, there is no significant increase in probe accuracy, suggesting that the model effectively captures emotional content by this stage. 

There is no clear distinction in probing performance between \( \mathbf{m}^{(l)} \) and \( \mathbf{h}^{(l)} \). Measurements from FFN closely track the hidden state dynamics, showing a steady increase in emotional conceptualization that peaks around the mid-layers. However, the heatmap corresponding to \( \mathbf{a}^{(l)} \) shows a more dispersed pattern while following the same consistent increasing trend observed in other locations. 
%The fluctuations observed even in the later layers suggest a specialization of attention units in each layer, with an increased potential for functional localization.

Our experiments reveal that emotion-processing mechanisms in LLMs are most pronounced in the middle layers across model families and sizes. \textit{These findings suggest that the model has largely determined the output emotion by the mid-layers, with subsequent layers adding little additional processing.} Our observation aligns with the understanding that higher transformer layers capture more abstract and task-specific features.

Lastly, to evaluate the hypothesis regarding the importance of the last token in causal modeling, we repeat the analysis on the last five tokens for Llama~3.2~1B in Appendix~\ref{app:token}. We observe a consistent increase in signal strength from earlier to later tokens, reinforcing the focus on the last token as the primary contributor to output generation.

\section{Emotion Transfer by Activation Patching} \label{sec:patching}

Given evidence suggesting that the model's internal representation of emotional content stabilizes around the mid-layers, we explore causal intervention in these regions to test their functional importance. Specifically, we assess whether the output emotion of a \textit{source} example can be transferred to a \textit{target} example, with a different emotion, by selectively patching activations from the source computation graph into the inference pass of the target at corresponding locations, a method referred to as activation patching \cite{patchscopes}. 

Formally, let \( \mathbf{x}^{(l)}_t \) be the activation vector from the target example, and \( \hat{\mathbf{x}}^{(l)}_t \) be the activations from a different example, i.e. the source sentence, which has a mismatched label from the target. The patching operation involves replacing the activation at layer \( l \) and token \( t \) by substituting \(\mathbf{x}^{(l)}_t \leftarrow \hat{\mathbf{x}}^{(l)}_t\) and letting the model continue the processing flow in the following layers and tokens.

The goal is to determine whether substituting specific activations with those from another example can manipulate the model's final prediction to reflect the intended label. We conduct experiments by substituting activations at the last token and within a window spanning five layers, consistently across all model families and sizes\footnote{Smaller models are more sensitive to fewer transferred layers, while larger models resist emotion transfer and require a larger span. For consistency, we use the same window size across all models.}. %For example, to test activation patching at the last token of layer 10, we replace the activations of the last token in layers \({l  \in [8-12]} \) from the source into the target activations and evaluate the new emotion labels.

The results shown in Figure~\ref{fig:e_patch_heat} demonstrate consistent behavior across model sizes and families. When intervening on hidden state activations, we observe a clear increase in intervention success (see Figure~\ref{fig:e_patch_heat} right). Take Llama~3.2~1B as an example. Patching hidden states in the early layers is entirely ineffective. There is a critical point, e.g. layer 10 in this model, after which copying the hidden state transfers the emotion label with a high success rate. A high success rate at the final layers is naturally an expected behavior, as the residual stream of hidden states aggregates the information as it gets closer to the final layers. However, remarkably, the chance of success peaks and stabilizes around the mid-layers in all models, an observation that aligns with our findings in the previous section.

To dig deeper, we look at the patching effect of MHSA and FFN units (see Figure~\ref{fig:e_patch_heat} left and middle, respectively), which provides clear evidence for functional localization of emotion processing. More precisely, interventions targeting \( \mathbf{a}^{(l)} \) and \( \mathbf{m}^{(l)} \) show clear evidence of success localized to specific mid-layers, e.g., MHSA units of layers \({l  \in [9-11]} \) in Llama~3.2~1B. In other words, successful patching of both MHSA and FFN units occurs only in a subset of layers and predominantly in the middle rather than the final layers, with the FFN's most successful patchings happening only slightly later than the MHSA's patching. \textit{We hypothesize that there are a few consecutive layers in each language model whose MHSA units are responsible for gathering emotional information from the rest of the tokens and integrating it into the hidden state of the last token. This mechanism is immediately followed by a processing in the subsequent FFN units.}

\begin{figure*}[t!] 
\centering
 \begin{subfigure}[b]{.8\textwidth}
    \centering
    \includegraphics[width=\linewidth]{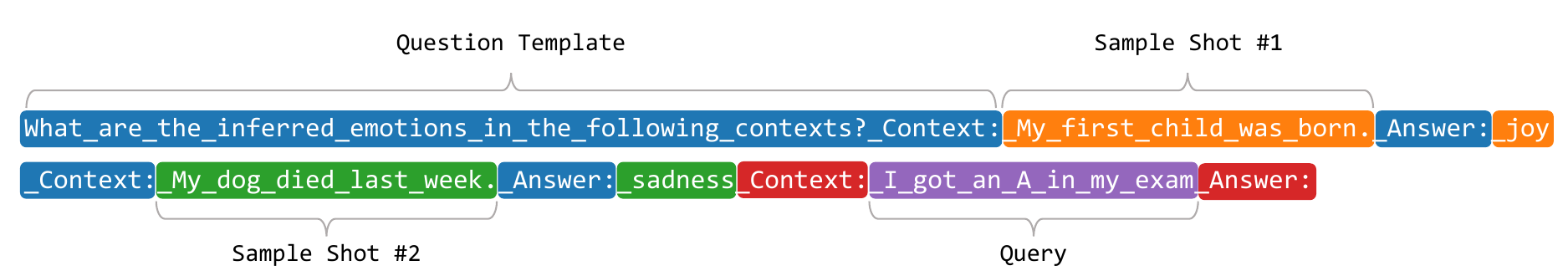}
\end{subfigure}
\hfill
\par\bigskip\bigskip
\centering
 \begin{subfigure}[b]{1.0\textwidth}
    \centering
    \includegraphics[width=\linewidth]{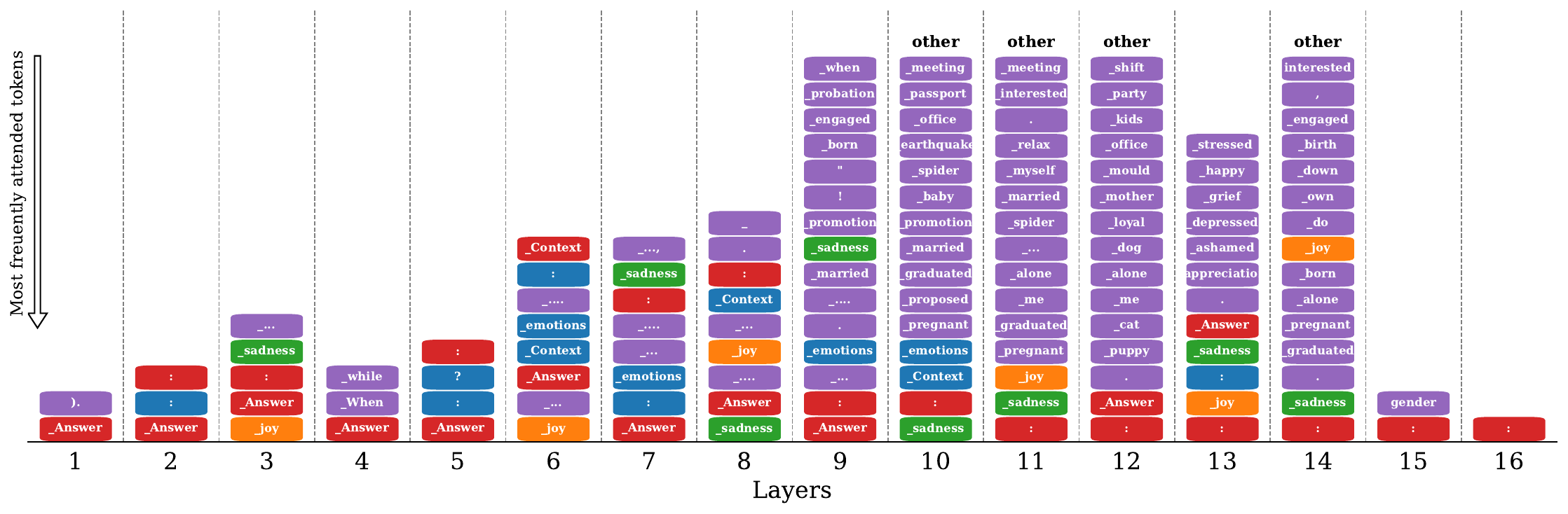}
\end{subfigure}
\caption {(\textbf{Top}) Primary prompt used in this study. Different segments of the prompt are colored differently. (\textbf{Bottom}) Most attended tokens at each layer from the perspective of the last token in Llama~3.2~1B. Layer 9 is the first layer in which the model attends to the tokens in the query with high emotional importance.}\label{fig:aggregate_attention_viz}
 \vspace{-2mm}
\end{figure*}

To complement these findings, we perform an additional experiment, where a set of activations is knocked out in the forward pass to assess their impact. The results in Appendix~\ref{app:zero} consistently align with those of our activation patching and probing, reinforcing the evidence of functional localization in emotion processing. These observations hold across different models and experimental methods.

Our results in all previous sections are not prompt-dependent. Specifically, changes in the format, wording, structure, or the number of demonstrations in the prompt do not affect the findings (see Appendix~\ref{app:prompt}). Additionally, we provide a control experiment in Appendix~\ref{app:control} to show that the identified units are not critical when performing a different yet syntactically similar task. In fact, the final layers are most critical for this isomorphic syntactic task, which differs significantly from the units we found for emotion processing.

%Additionally, we explore whether the observed outcomes could be attributed solely to syntactic features and task structure by conducting an isomorphic experiment. We modified the task to focus purely on syntax—predicting the first word in the sequence—and repeated the activation patching procedure. As detailed in Appendix~\ref{app:control}, the final layers of the model are most critical for this syntactic task, which contrasts significantly with the emotion patching findings.

%\subsection{Attention Visualization}

To further explain our findings, we analyze the attention patterns in Llama~3.2~1B. Specifically, we record the top 3 tokens attended to by all attention heads in the last token of each layer. We conduct this analysis for all samples in the dataset, providing insight into which tokens are most frequently attended to. %For clarity, we categorize the prompt into multiple segments: initial instructions, demonstration samples, the query, and 
%lastly the pre- and post-syntax words, e.g., \textit{"\_Context"}, \textit{"\_Answer"}, and \textit{":"}.
%(See Figure~\ref{fig:aggregate_attention_viz} top).
Figure~\ref{fig:aggregate_attention_viz} presents the aggregate attention patterns. The results indicate that early layers primarily focus on syntax. Around the mid-layers, the model shifts its attention to emotionally relevant tokens in the query and maintains this focus until the final layers. In the last few layers, attention predominantly focuses on the last token in the sequence, suggesting that it primarily carries forward the last token's hidden state from earlier layers. These patterns provide evidence that the identified middle units are meaningfully related to emotion processing.

% However, when no examples are provided (zero-shot), the model's behavior deviates significantly. These findings suggest that the analyses and results in this study are primarily applicable to scenarios involving few-shot prompts. This raises the question of whether the identified regions are more fundamental than emotion processing itself, potentially serving as high-level enablers for emotion signals or the causal mechanisms behind emotion prediction (as demonstrated in our experiments).

\section{Investigating Appraisal Concepts}
\label{sec:appraisal_modulation}

We draw inspiration from cognitive appraisal theory to show an existing emotion structure in LLMs' latent representations. Appraisals are known to have significant associations with emotions and, in some accounts, are considered causal factors in their emergence \cite{rosenman2001appraisal}.

\begin{figure}[t]
    \centering
  \includegraphics[width=0.75\linewidth]{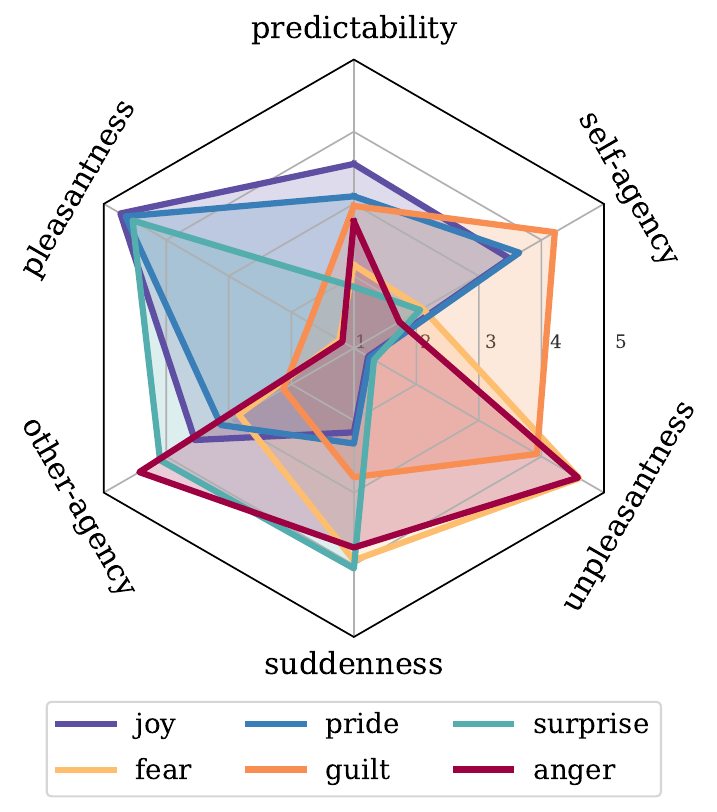} 
  \caption {Appraisal emotion associations extracted from the dataset. For example, when participants report \textit{anger}, they typically perceive a high degree of \textit{other-agency} and a low level of \textit{pleasantness} in the situation.}
  \label{fig:app_emo_cor}
  \vspace{-2mm}
\end{figure}

Figure~\ref{fig:app_emo_cor} illustrates three primary appraisal dimensions, including \textit{pleasantness}, \textit{other-agency}, and \textit{predictability}, in addition to their associations with a set of basic emotions. For instance, \textit{guilt} and \textit{pride} are both materialized in situations with high \textit{self-agency}, with the former happening in \textit{pleasant} situations while the latter comes with an \textit{unpleasant} experience. These mappings, which align closely with prior findings in appraisal theory \cite{Ellsworth2015}, are extracted from our dataset by averaging appraisal scores for each emotion label.

We begin by training linear probes for appraisal signals within the model representations. In contrast to the linear classification probes used in Section~\ref{sec:emotion_probe}, here we solve multiple independent regression tasks for each appraisal. More formally, consider a set of $n$ appraisals and let $\mathbf{v}_a \in \mathbb{R}^d$ represent the weight vector corresponding to the appraisal $a \in \{1,\cdots,n\}$. Let $\mathbf{x}$ be an activation vector, as introduced in Section~\ref{sec:notations}. We train the regression weight $\mathbf{v}_a$ and the bias $b_a \in \mathbb{R}$ such that,

\[
    \hat{r}_a = \mathbf{v}_a^\top \mathbf{x} + b_a,
\]

where $\hat{r}_a \in \mathbb{R}$ is the estimate for the appraisal score $a$, given the input $\mathbf{x}$.
We train a separate appraisal probe for each appraisal at the hidden states of each layer. The weight matrices obtained through this process serve as representations of the corresponding appraisal, encoding features of the appraisal concept at the activations of the specified locations. The success of appraisal probing highly depends on whether the target concept is linearly detectable in the targeted activation. We provide the appraisal probing results in Appendix~\ref{app:appraisal_probing}, showing that the appraisal signals are not linearly detectable at earlier layers but are strongly present as we approach the hidden state of the final layers.

\section{Emotion-Appraisal Mappings}
\label{sec:app_emo_projection}
In this section, we analyze the representations of emotions and appraisals to reveal a structure within the latent LLM representations. Remember the weight matrix $\mathbf{W} \in \mathbb{R}^{d\times C}$ introduced in Section~\ref{sec:emotion_probe}. Let $\mathbf{w}_e \in \mathbb{R}^d$ represent the column $e$ of $\mathbf{W}$ corresponding to the emotion index $e \in \{1, \cdots, C\}$. We define the cosine similarity of appraisal $a$ with emotion $e$ as $\textit{sim($a$,$e$)}=\frac{\mathbf{v}^\top_a \mathbf{w}_e}{\norm{\mathbf{v}_a}_2 \norm{\mathbf{w}_e}_2}$.

\begin{figure}[t]
    % \centering
\includegraphics[width=0.85\columnwidth]{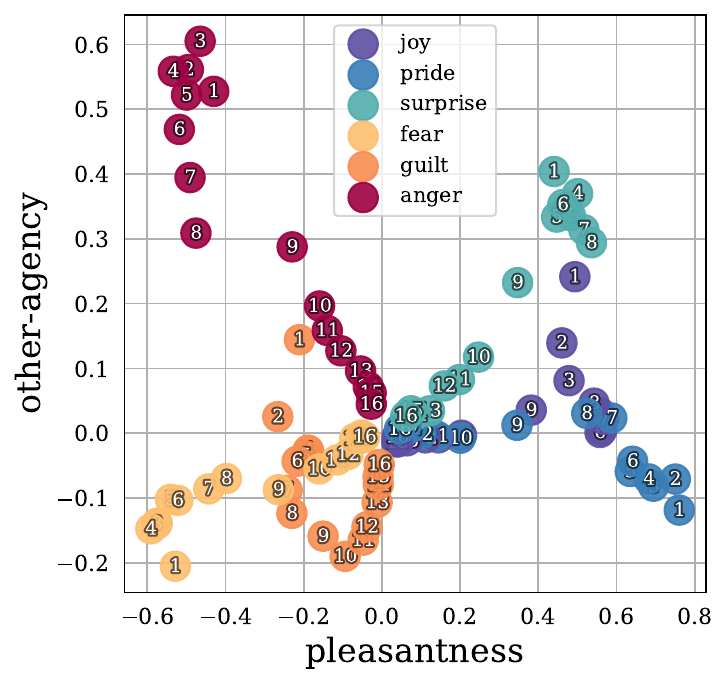} 
  \caption {Cosine similarity of emotion vectors with \textit{pleasantness} and \textit{other-agency} appraisal vectors extracted from the hidden state of different Llama~3.2~1B layers. Layer number is written inside each marker.}
  \label{fig:emo_app_projection}
  \vspace{-2mm}
\end{figure}

Figure~\ref{fig:emo_app_projection} shows the similarity score of emotion vectors with two appraisal vectors, i.e. the \textit{pleasantness} and \textit{other-agency}, throughout Llama~3.2~1B layers. Notably, we observe psychologically plausible appraisal-emotion mappings across all layers. However, the projection strength peaks in the early layers and fades to near zero in the final layer, suggesting orthogonality in the final layers. %This observation is counter-intuitive to our earlier findings that emotion and appraisal signals are mostly linearly identifiable in the later layers. 

We hypothesize that in the earlier layers, there exists a meaningful structure capturing appraisal and emotion concepts, which aligns with our expectations from the appraisal theory. However, these concepts gradually decouple as the processing progresses through the network, and by the final layers, they become orthogonal and fully decoupled, reflecting the specialization of the network toward higher-level tasks. We finish this section by drawing the connection to our findings in previous sections. Notice that the decoupling starts around the critical layers, e.g. layers 9 and 10 in Llama~3.2~1B, which we identified in previous sections. Therefore, our observations suggest that \textit{the appraisals build a foundation to understand emotion representations in LLM hidden states, but the structure vanishes as we progress through the network.}

\section{Intervention on Appraisal Concepts} \label{sec:appraisal_intervention}

After finding the appraisal vectors, we investigate the possibility of indirectly modifying the emotion of an input example by modulating its appraisals within the model representations. For this purpose, we need to isolate the role of each appraisal \( a \), by considering its associated latent vector \( \mathbf{v}_a \) and distinguishing it from other appraisal vectors. More precisely, we define \({
    \mathbf{V}_{-a} := \left[ \mathbf{v}_1, \dots, \mathbf{v}_{a-1}, \mathbf{v}_{a+1}, \dots, \mathbf{v}_n \right]}
\) by concatenating all appraisal vectors except \( \mathbf{v}_a \). Next, we introduce the \textit{unique effect vector} of appraisal $a$ 
as ${\mathbf{z}_a := (\mathbf{I} - \mathbf{P}_{-a}) \mathbf{v}_a}$, where \( {\mathbf{P}_{-a} = \mathbf{V}_{-a} (\mathbf{V}^\top_{-a}\mathbf{V}_{-a})^{-1} \mathbf{V}^\top_{-a}}\) is the projection matrix onto the column space of \( \mathbf{V}_{-a} \) and ${\mathbf{I}\in \mathbb{R}^{d \times d}}$ is the identity matrix. We perform \textit{appraisal modulation} by injecting \( \mathbf{z}_a \) into the model's latent representation. More formally, the intervention is expressed as:

\vspace{-2mm}
\[
    \mathbf{x} \leftarrow \mathbf{x} + \beta \frac{\mathbf{z}_a}{\norm{\mathbf{z}_a}_2},
\]

where \( \mathbf{x} \) on the RHS is the original latent representation, e.g., a hidden state vector from a specific layer, and \( \beta \) is a scaling factor controlling the strength of the concept modulation. Notice that a positive $\beta$ corresponds to an appraisal promotion while a negative $\beta$ has the opposite effect of appraisal demotion. To measure the success of interventions, we evaluate the new emotion label derived from this modification and repeat this procedure across all examples. Notice that, this modification is applicable to any layer of the model.

\begin{figure*}[ht!]
  % \includegraphics[width=.9\linewidth]
  % {figs/app_modulation.png} \vfill
  \centering 
  \includegraphics[width=1.0\linewidth]
  {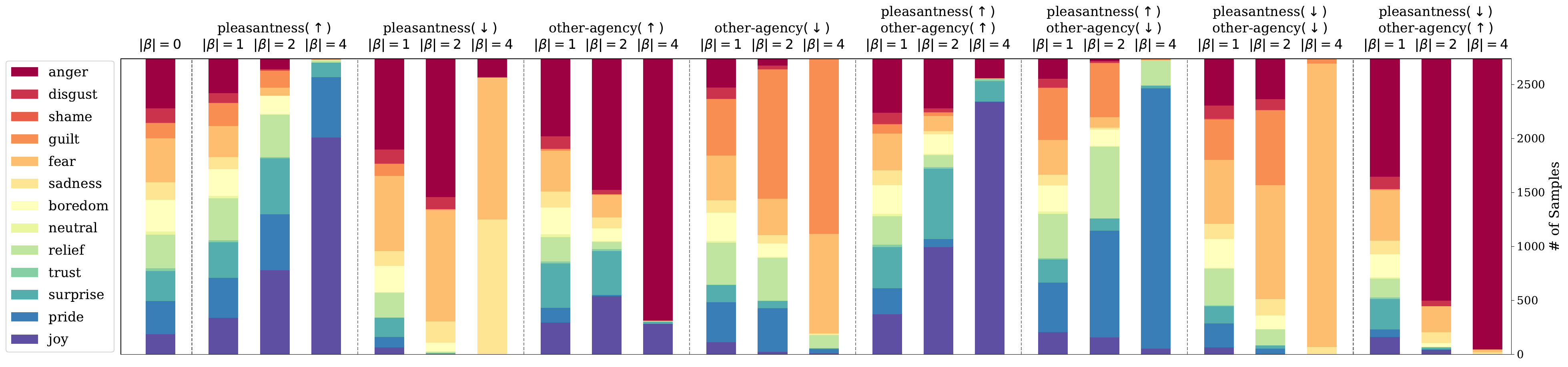}
  \caption {
Results of appraisal concept modulation by intervening at layer~9 hidden states of Llama~3.2~1B for increasing scaling factors (\(\beta\)). \(|\beta| = 0\) represents the original distribution of emotions in the dataset. For example, promoting ($\uparrow$) or demoting ($\downarrow$) \textit{other-agency} significantly increases the share of \textit{anger} and \textit{guilt}, respectively. Similarly, promoting or demoting \textit{pleasantness} increases the share of \textit{joy}/\textit{pride} and \textit{sadness}/\textit{guilt}/\textit{anger} outputs, respectively. Additionally, promoting \textit{pleasant other-agency} significantly increases the share of \textit{joy} outputs, while the promotion of \textit{unpleasant other-agency} significantly increases the share of \textit{anger}.
  }
  \label{fig:app_modulation2}
  \vspace{-2mm}
\end{figure*}

Figure~\ref{fig:app_modulation2} illustrates concept modulation results with different magnitudes of $\beta$ on layer $9$ of Llama~3.2~1B. Notably, we observe a remarkable alignment with theoretical and intuitive expectations. For instance, increasing the \textit{pleasantness} appraisal promotes both \textit{joy} and \textit{pride}, aligning with the fact that both of these emotions have high associations with the \textit{pleasantness} appraisal. 

In contrast to these results, the appraisal modulation when applied to earlier layers, does not generate psychologically valid results and is totally ineffective when applied to later layers. This observation matches the intuitions on the mechanism we provided earlier; Intervening on the early layers is not valid since modifications to latent representations are overwritten by emotion-specialized mid-layers. On the other hand, intervention on final layers is not effective because of the orthogonality of concepts as demonstrated in Section \ref{sec:app_emo_projection}. Appendix~\ref{app:appraisal_modulation} presents the full results, including intervention across all layers of Llama~3.2~1B.

Appraisal theory is also predictive of the situations in which two appraisals are promoted simultaneously. To test this capability in LLMs, we perform an intervention on the superposition of two appraisal dimensions: \textit{other-agency} and \textit{pleasantness}, with mathematical details provided in Appendix~\ref{app:appraisal_modulation}. The results, depicted in Figure~\ref{fig:app_modulation2}, show a successful promotion of emotion \textit{pride} with no further occurrences of \textit{joy}. These findings provide strong evidence that layer 9 in Llama~3.2~1B directly contributes to cognitive processes related to emotions (see Appendix~\ref{app:appraisal_modulation} for experiments with other appraisal concepts). Additionally, we provide complementary experiments such as direct emotion promotion using emotion vectors in Appendix~\ref{app:emotion_promotion}.

\section{Discussion}
\label{sec:discussion}
We employed mechanistic interpretability techniques to investigate the inner workings of emotion inference in LLMs. Our results reveal that mid-layer MHSA units within these models are responsible for processing emotional content. By applying linear algebraic manipulations to modulate the antecedents of emotions, i.e. the appraisal concepts, we steered the model outputs in controlled and predictable ways. This is particularly important for ensuring the \textit{reliability and steerability} of LLMs in high-stakes affective domains such as legal decision-making and clinical therapy. 

A key distinction of our work is that we grounded our experiments on psychological theory and applied MI analysis on \textit{in-the-wild examples}, rather than relying on \textit{synthetically generated simplistic structures}, as seen in prior studies \cite{ 2024circuitreuse}. For example, \citet{wang2022interpretability} study the Indirect Object Identification task, by considering a fixed input structure, such as \textit{``person1 and person2 had fun at school. person2 gave a ring to”} where the model is expected to predict \textit{``person1”}. \citet{hanna2023doesgpt2} study the ``Greater-than" task using a dataset of examples like \textit{``The war lasted from 1517 to 15”}, where the model is expected to predict any two-digit number larger than 17. While such notable efforts have demonstrated the feasibility of identifying end-to-end specialized circuits for these tasks, effectively interpreting them still requires structured inputs, making it challenging to generalize findings to more naturalistic settings.

Our work also highlights new opportunities for future research. Despite significant advancements in understanding human emotions, debates persist regarding the definition of emotion, the role of cognition in emotion, and the mechanisms underlying emotion inference \cite{ortony2022cognitive, ellsworth2003appraisal, moors2013causal, barrett2017theory}. In parallel, cognitive neuroscience has explored the neural basis of emotion in support of differing theoretical perspectives \cite{kragel2024can}. The study of LLMs, combined with insights from emotion theory and neuroscience, opens a unique intersection for advancing our understanding of emotions \cite{thornton_neuroscience}. 

Furthermore, our steering approach opens promising possibilities for conditioning LLMs to exhibit specific personality traits or moods, which could benefit applications requiring tailored affective responses \cite{jiang2023personality, jiang268032940personallm, petrov2024limited, suh2024rediscovering, li2024big5, suh2024rediscovering}. However, to ensure these interventions do not introduce unintended disruptions to other critical language-processing functions, it is essential to rigorously evaluate models on standard NLP benchmarks after inducing traits or moods.

Given LLMs’ increasing societal impact —spanning areas such as mental health, legal decision-making, and human-AI interaction—it is imperative to deepen our understanding of their internal mechanisms. Our study breaks new ground in the interpretability of emotion inference in LLMs, offering a novel way to causally intervene in emotional text generation. These findings hold promise for improving safety and alignment in sensitive affective domains. Moving beyond black-box approaches to rigorously test and refine LLM emotion processing will not only advance the field of LLM interpretability but also unlock new pathways for more responsible AI systems.

\section{Limitations}
\label{sec:Limitations}

In this study, we build upon the linear representation hypothesis \cite{mikolov2013exploiting,mikolov-etal-2013-linguistic, levy2014linguistic, elhage2022toymodelssuperposition}—the idea that high-level concepts are encoded linearly within model representations \cite{parklinearhype}. This hypothesis is particularly appealing because, if true, it could enable simple and effective methods for interpreting and controlling LLMs—an approach we leveraged to localize and manipulate latent emotion representations. However, despite notable efforts to formalize the notions of linearity \cite{parklinearhype} and orthogonality \cite{jiang2024uncovering} in model representations, recent research suggests that not all features are encoded linearly \cite{engels2025not}. Further investigation is needed to improve clarity and robustness in this area.

Furthermore, we demonstrated the ability to manipulate affective outputs by modifying appraisal concepts. Nevertheless, the precise nature of this relationship remains unclear—it is possible that appraisals are merely correlated with emotions rather than exerting a direct causal influence or that the relationship follows an inverse causal pattern. Establishing causality requires further investigation in future studies to disentangle directional dependencies. A deeper understanding of the interplay between LLM emotional inference, emotion theory, and neuroscience will be crucial for both theoretical insights and practical applications. Addressing these challenges will refine our understanding of LLMs and enhance their reliability in affective computing.

\section{Ethical Impact Statement}
\label{sec:ethic}
This study re-analyzes previously collected, de-identified data that had already undergone ethical review. The dataset is used for investigating the inner mechanisms by which auto-regressive LLMs process emotion. However, caution must be exercised when generalizing these results to models not examined in this work, to superficially similar tasks, or to different languages. 

Our analysis highlights potential concerns for those deploying LLMs in high-stakes affective domains or for generating emotionally charged content. Given the risks associated with emotional manipulation by LLMs, it is crucial to develop a deeper understanding of how these models process emotions. To this end, we advocate for further research in this domain to ensure that LLMs align with ethical standards and human-centered AI principles.

\section*{Acknowledgments}
This work is, in part, supported by the Army Research Office under Cooperative Agreement Number W911NF-25-2-0040. Only staff at ICT were sponsored directly by the Army Research Office. The views and conclusions contained in this document are those of the authors and should not be interpreted as representing the official policies, either expressed or implied, of the Army Research Office or the U.S. Government. The U.S. Government is authorized to reproduce and distribute reprints for Government purposes notwithstanding any copyright notation herein.

\bibliography{main}

\clearpage 

\appendix
\section{Experimental Details}

\subsection{Dataset Details} \label{app:dataset}
For this project, we employed the crowd-enVENT dataset. \citet{envent} developed the dataset by asking crowdsource writers to share an event that made them feel a particular emotion. The vignettes varied in length, ranging from short sentences to longer narratives, but the language was predominantly everyday English. The participants were then asked to evaluate their subjective experiences during that event, including their perceived appraisals. Both were rated on a 5-point Likert scale with 5 representing the highest agreement. 
%Finally, their written account of the event was shown to annotators who were tasked with identifying the emotion of the writer and their appraisals of the event.

The dataset is comprised of examples from the following list of emotions: \textit{Joy, Pride, Surprise, Trust, Relief, Neutral, Boredom, Sadness, Fear, Guilt, Shame, Disgust, and Anger}. The dataset has 500 examples for each emotion, except for guilt and shame, which have 250 samples. The events were appraised along the following dimensions:  
\textit{pleasantness, other-agency, predictability, suddenness, familiarity, unpleasantness, goal-relatedness, 
own responsibility,  situational responsibility, goal support,
consequence anticipation, urgency of response, own control, others’ control, situational control, accepted control, internal standards, external norms, attention, not considered, and effort}. We selected a subset of dimensions for our analyses, previously shown to have a high association with emotions \cite{Ellsworth2015, tak2024gpt4emulatesaveragehumanemotional}.

To give a few examples from the dataset, the emotion label for the sentence ``I baked a delicious strawberry cobbler'' is \textit{pride}, with appraisals $\text{\textit{pleasantness}}=5$, $\text{\textit{other-agency}}=1$, while ``A housemate came at me with a knife'' is an example of \textit{fear} with $\text{\textit{pleasantness}}=1$ and $\text{\textit{other-agency}}=5$.

\begin{table*}[ht]
\centering
\tiny
\caption{Architectural details of the language models used in this study.}
\begin{tabular}{ cccccc }
\hline
 & Llama~3.2~1B Instruct& Llama~3.1~8B Instruct& Gemma~2~2b-it& Gemma2~9b-it &Ministral~8B Instruct
 \\
\hline
Parameters & 1B & 8B & 2B & 9B  &8B\\  
hidden size $d$ & 2048& 4096& 2304& 3584& 4096\\  
Layers & 16& 32& 26& 42&36\\   
layer norm type & RMSNorm & RMSNorm & RMSNorm & RMSNorm & RMSNorm\\  
Non-linearity & SiLU& SiLU& GeLU& GeLU&SiLU\\  
Feedforward dim & 8192& 14336& 9216& 14336&12288\\  
Head type & GQA  & GQA  & GQA  & GQA &GQA \\  
Num heads & 32& 32& 8& 16&32\\  
Num KV heads & 8& 8& 4& 8 & 8\\    
Context Window& 131072& 131072& 8192& 8192&32768\\  
Vocab size & 128256& 128256& 256000& 256000&131072\\
Tied embedding & True& False& True& True&False\\
\hline 
\end{tabular}
\newline
\newline
\newline
\newline
\newline
\centering
\begin{tabular}{ cccccc }
\hline
 & Mistral~12B Nemo Instruct& Phi~3.5 mini Instruct& Phi~3~medium Instruct& OLMo~2~7B Instruct & OLMo~2~13B Instruct\\
\hline
Parameters & 12.2B& 3B & 14B& 7B& 13B\\  
hidden size $d$ & 5120& 3072 & 5120& 4096& 5120\\  
Layers & 40& 32 & 40& 32& 40\\   
layer norm type & RMSNorm & RMSNorm & RMSNorm & RMSNorm& RMSNorm\\  
Non-linearity & SiLU& SiLU& SiLU& SiLU& SiLU\\  
Feedforward dim & 14336& 8192&  17920& 11008& 13824\\  
Head type & GQA & Multi-Head & GQA & Multi-Head & Multi-Head \\  
Num heads & 32& 32 & 40& 32 & 40\\  
Num KV heads & 8& 32& 10& 32 & 40\\   
Context Window & 131072& 131072& 131072& 4096& 4096\\  
Vocab size & 131072& 32064& 32064& 100352& 100352\\
Tied embedding & False& False& False& False& False\\
\hline 
\end{tabular}
\label{tab:model_arch}
\end{table*}

\subsection{Architecture and Model Details} \label{app:arch}

In this paper we experimented with ten LLMs: meta-llama/Llama-3.2-1B-Instruct and meta-llama/Llama-3.1-8B-Instruct \cite{llama3}, google/gemma-2b-it and google/gemma-2-9b-it \cite{gemma2}, allenai/OLMo-7B-Instruct and allenai/OLMo-2-1124-13B-Instruct \cite{olmo2}, microsoft/Phi-3.5-mini-instruct and microsoft/Phi-3-medium-128k-instruct \cite{abdin2024phi}, and mistralai/Ministral-8B-Instruct-2410 and nvidia/Mistral-NeMo-12B-Instruct \cite{mistral2024nemo}.  The architectural details for these language models are provided in Table~\ref{tab:model_arch}. Unless stated otherwise, the default model used in this paper is Llama~3.2~1B since it is the lightest model, allowing for efficient analysis.

All models were implemented using the \href{https://huggingface.co/models}{Hugging Face framework}\footnote{\url{https://huggingface.co/models}}, leveraging the respective model weights, with additional integration of libraries such as TransformerLens \cite{nandatransformerlens2022} to enable the hooking and intervention on hidden states and activations.

\begin{figure}[ht]
    \centering
    \includegraphics[width=1.0\columnwidth]{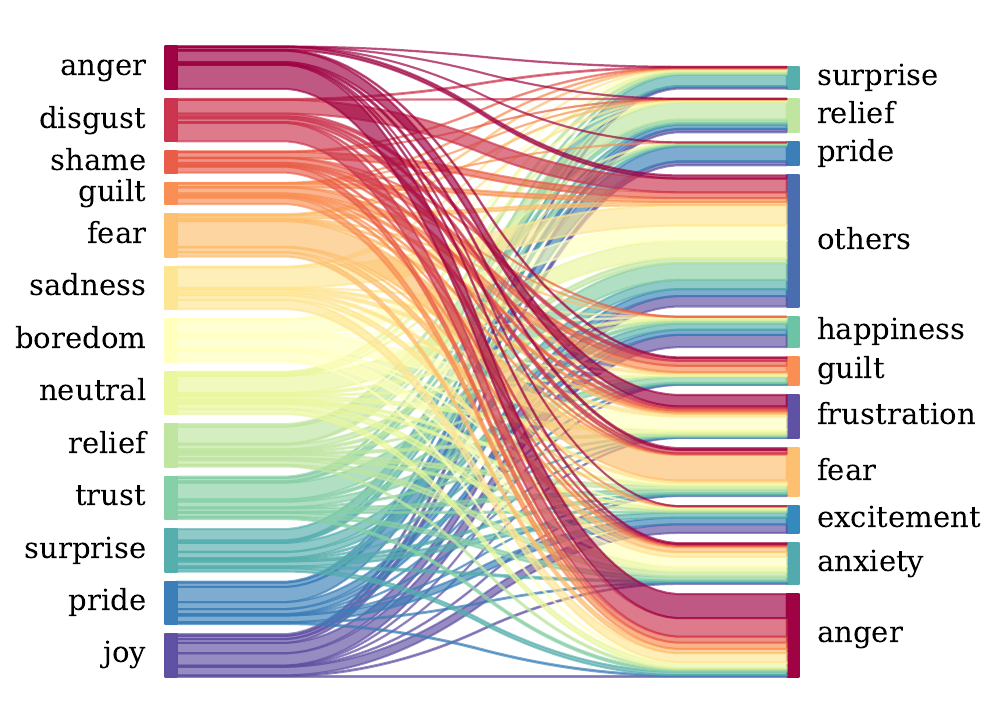}
  \caption{Llama~3.2~1B open vocab generation of emotions, comparing the true label to the predicted label through open vocab prediction}
  \label{fig:Llama1B_open_vocab}
\end{figure} 

\begin{figure}[ht]
    \centering
    \includegraphics[width=0.9\columnwidth]{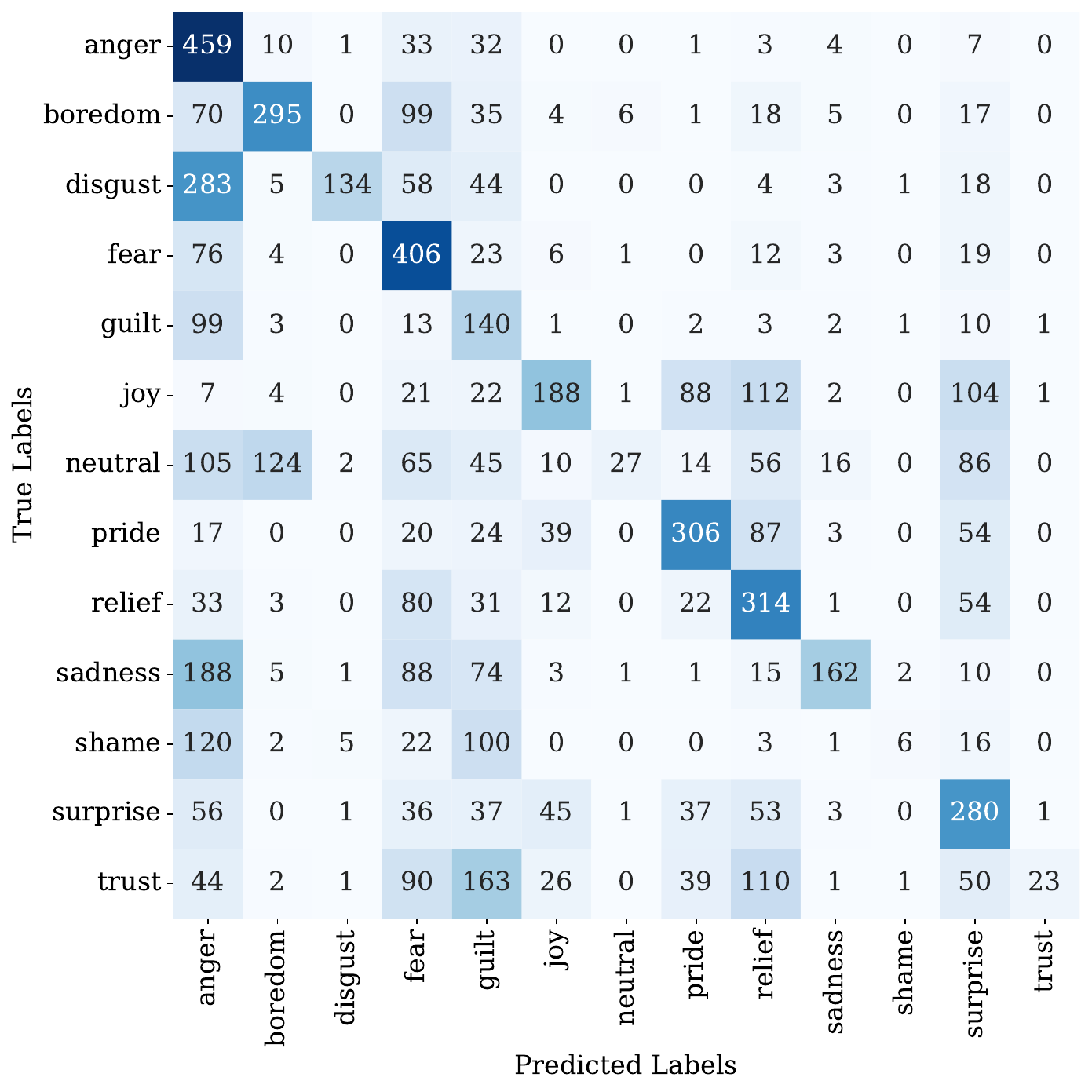}
  \caption{Confusion matrix comparing the true labels to Llama~3.2~1B predicted labels}
  \label{fig:Llama1B_confusion_matrix}
\end{figure} 

\begin{figure*}[t]
    \centering
    \includegraphics[width=0.8\textwidth]{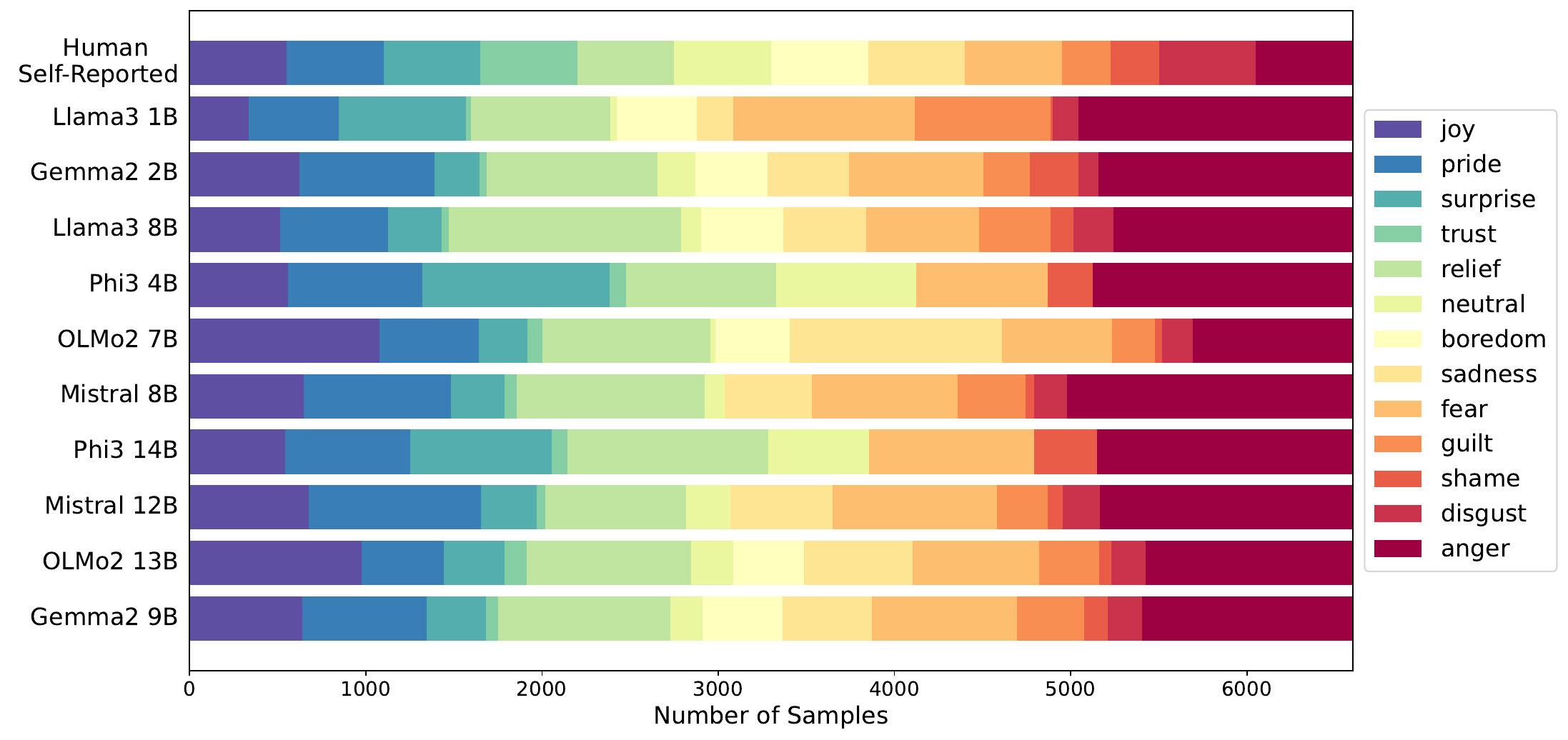}
  \caption{The distribution of next word emotion label predictions. The self-reported human labels are shown on the top row. The subsequent rows demonstrate the distribution of the predicted emotion labels from each model.}
  \label{fig:models_pred_distribution}
\end{figure*} 

\begin{figure}[t]
    \centering
    \includegraphics[width=1.0\columnwidth]{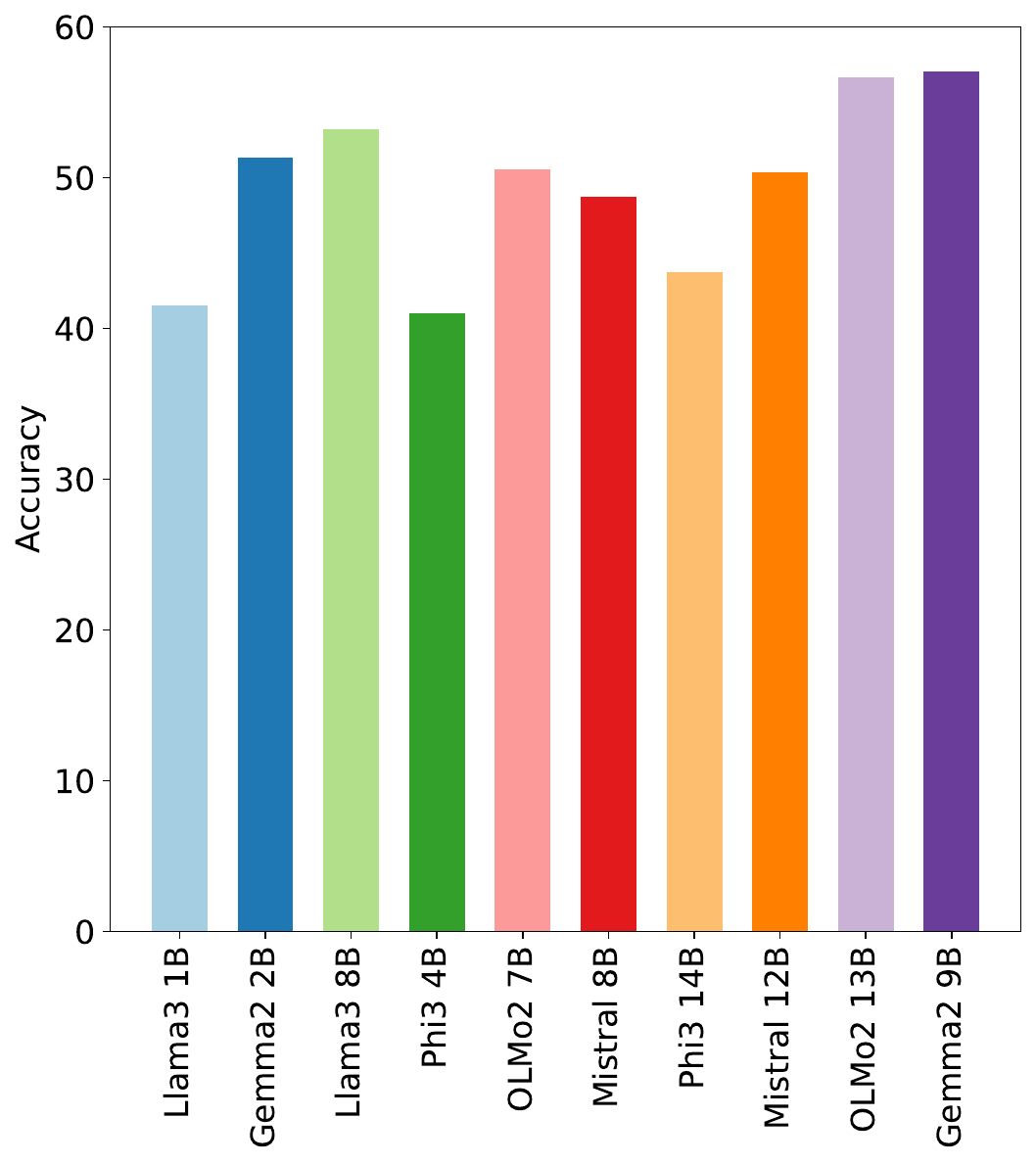}
  \caption{Comparison of models' accuracy on the emotion classification task. This experiment is performed across all samples in the dataset.}
  \label{fig:models_accuracy}
\end{figure} 

\subsection{Task Details} \label{app:task}

Emotion attribution is a challenging and subjective task, one that even humans often find difficult. The accuracy of third-person human annotations is approximately 50\% in this dataset, as calculated using 1000 bootstrap resampling with 95\% confidence intervals (the chance baseline being around 7\%) \cite{envent}. Thus, it is not a surprise that defining a ground truth emotion label for each sample is challenging. Consequently, it is reasonable to expect that this task would also be challenging for LLMs. Figure \ref{fig:Llama1B_open_vocab} demonstrates the emotion label prediction on an open vocabulary, for Llama~3.2~1B. As the figure suggests, the LLM vocab choice in the output rarely matches the human-reported emotion label.

Therefore, we avoid doing an open-vocab generation, but instead, we confine the logits of LLM output to the set of emotion labels in the dataset. With this modification, we achieve an accuracy of approximately 40\% or higher using self-reported ground truth emotions across all architectures and scales. Figure \ref{fig:Llama1B_confusion_matrix} demonstrates the closed vocabulary results through a confusion matrix for Llama~3.2~1B, comparing the true and predicted labels. Furthermore, Figure~\ref{fig:models_pred_distribution} illustrates the emotion label predictions across the LLMs tested, and Figure~\ref{fig:models_accuracy} demonstrates the accuracy results from these experiments.

For the rest of our analysis, we only focus on the correctly classified examples, which ensures at least 2,700 data points or more across different model architectures. We apply this filtering method to ensure that the samples selected for experimentation are ones where the LLM understood the emotion labeling task, allowing us to properly investigate the underlying mechanisms that led to the models' selected emotion label.  
However, we acknowledge that some emotion classes have a limited number of examples after filtering, which may present constraints in our experiments. A preliminary study targeting the bias of such filtering is provided in Appendix~\ref{app:mistakes}, while deeper investigations are left for future studies.

Most experiments in this paper did not have a stochastic nature or behavior, and we used all the available data. The only source of stochasticity in our experiments comes from the train/test split in probe training. However, after extensive experiments, we noticed that the variance for these experiments is extremely low. More precisely, we observe no more than 1.5\% STD in classification accuracy across 5 different seeds of each individual probe across all layers, with $R^2$ differing by no more than 0.02 between runs for regression tasks. The reported values in the paper figures are therefore only the mean values. Note that we used 5-fold cross-validation for probe hyperparameter tuning; therefore, no extra validation data split was necessary.

\section{Knockout Experiment} \label{app:zero}
In this section, we elaborate on a causal approach in MI commonly referred to as \textit{ablation}, \textit{knockout}, or \textit{zero/random activation intervention} \cite{rai2024practical, olsson2022context, wang2022interpretability,chen2023sudden}. This is a complementary approach to the activation patching experiment to provide further evidence for the localization of emotion processing in LLMs. Precisely, we zero out the activations at different points in the model or replace them with random activations and assess the impact on the generated output label. In other words, we intervene in the activations in the forward pass of the model at MHSA, FFN, and the hidden states at the last token across different layers.
Formally, let \( \mathbf{x}\) be the activation vector from the target example at a specific layer and location. The zero-activation operation involves replacing the activation by substituting \(\mathbf{x} \leftarrow \mathbf{0}\) and letting the model continue the processing flow in the following layers. Similarly, for random intervention, let \( \mathbf{x} \) be the target activation to interrupt. The random activation intervention replaces \( \mathbf{x} \) by substituting
\[
\mathbf{x} \leftarrow \frac{\norm{\mathbf{x}}}{\norm{\mathbf{r}}}\mathbf{r},
\]

where \( \mathbf{r} \in \mathbb{R}^d \) is sampled from a standard Gaussian $\mathcal{N}(\mathbf{0}, \mathbf{I})$ distribution and the normalization factor $\frac{\norm{\mathbf{x}}}{\norm{\mathbf{r}}}$ ensures that the new activation has the same norm as the original one.

The modified activations propagate forward, affecting the model's outputs. Then, we compare the prediction coming out of the modified logits with the clean forward pass to measure the model's accuracy after the intervention. The lower this accuracy is, the more significant that activation is affecting emotion label generation.

\begin{sidewaysfigure*}[bht]
    \centering
    \includegraphics[width=0.99\textwidth]{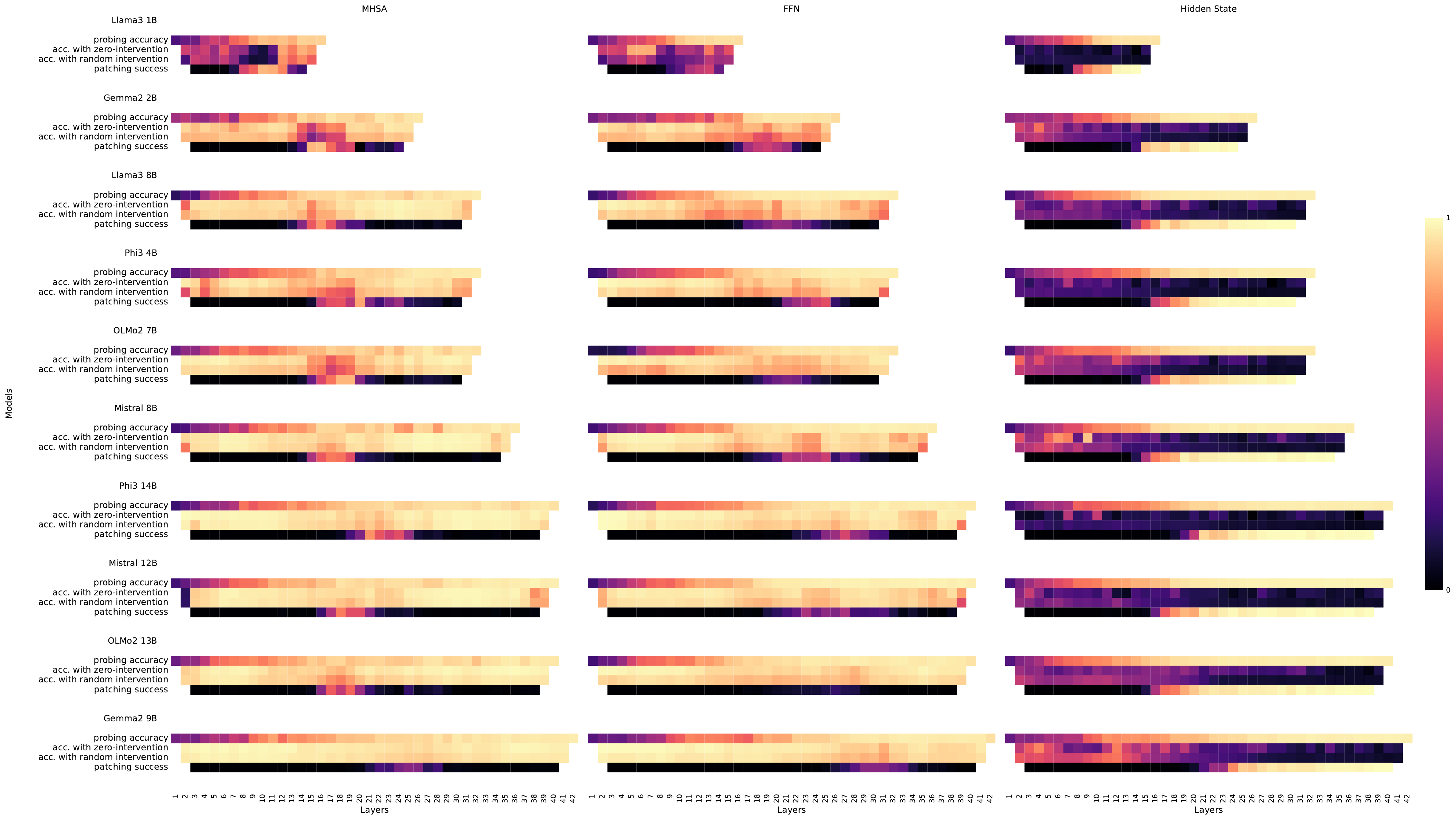}
    \caption{
    Comparison of probing, zero-activation and random-activation interventions, and activation patching on MSHA, FFN, and hidden state units across all layers of Llama~3.2~1B.  
    The probe heatmap shows accuracy on the holdout set, zero/random activation interventions measure the model accuracy after disrupting causal pathways, and patching heatmaps indicate how effectively outputs transfer from source to target examples. The span sizes of $3$ and $5$ are used for the presented knockout interventions and patching experiments. This means that we intervene simultaneously on three/five consecutive layers, with the center being the indicated layer.
    }
    \label{fig:zero_all}
\end{sidewaysfigure*}

In Figure~\ref{fig:zero_all}, knockout-intervention across all models, we find remarkably consistent behavior with our probing and patching results provided in Sections~\ref{sec:emotion_probe} and ~\ref{sec:patching}—that after a certain point, even removing all MHSA units and, to some extent, the FFN units do not impact the final emotion classification accuracy. This indicates that the model's internal representation of the emotional content is established before that point. Additionally, we observe that knocking out activations with both zero and random interventions at \( \mathbf{a}^{(10)} \) has a significantly greater impact than \(\mathbf{m}^{(10)} \), which suggests that the MSHA unit in mid layers plays a more crucial role in collecting the emotion label from previous tokens.

For example, given the first row in the plot, we observe that zeroing out \( \mathbf{a}^{(9-11)} \) and \( \mathbf{m}^{(9-11)}\) LLama~3.2~1B has the greatest impact on the emotion label (corresponding to each example). As expected, activations at \( \mathbf{h}^{(l)} \) have a significant impact throughout all layers since they constitute the mainstream of the forward path reaching the model's output and are fundamentally different from \( \mathbf{a}^{(l)} \) and \( \mathbf{m}^{(l)} \), which contribute additional processing to the residual stream. 

\clearpage
\clearpage

\clearpage
\section{Case Studies on Llama~3.2~1B}
In this section, we focus on the Llama~3.2~1B language model and investigate the validity of our findings from multiple aspects. In Section~\ref{app:detailed_figures}, we provide detailed results on emotion probing, activation patching and knockout interventions. In Section~\ref{app:token} we extend our studies to include token dimension. Finally, we show that our results are robust to prompt design by conducting experiments using several hand-designed prompts in Section~\ref{app:prompt}.

\begin{figure}[bht]
    \centering
    \includegraphics[width=1.0\columnwidth]{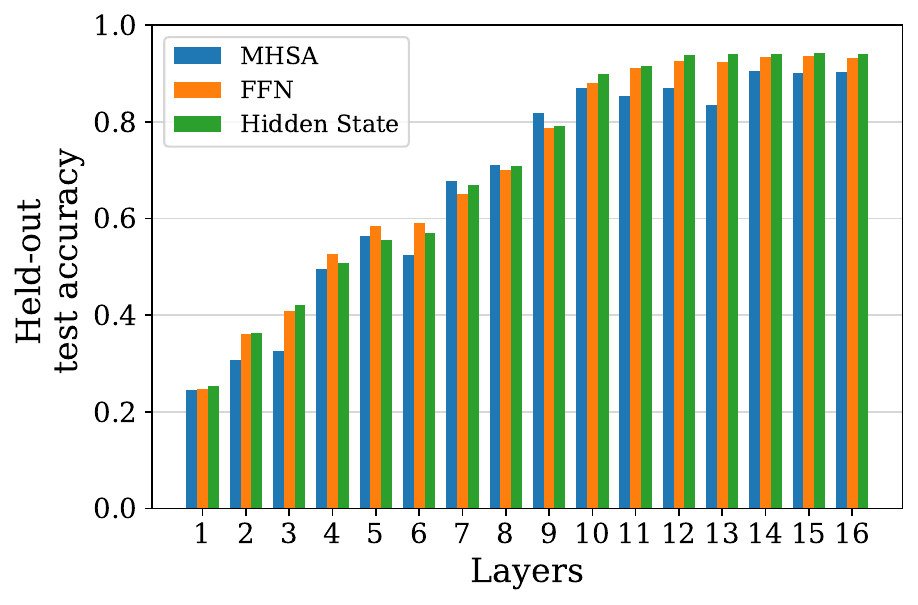}

    \includegraphics[width=1.0\columnwidth]{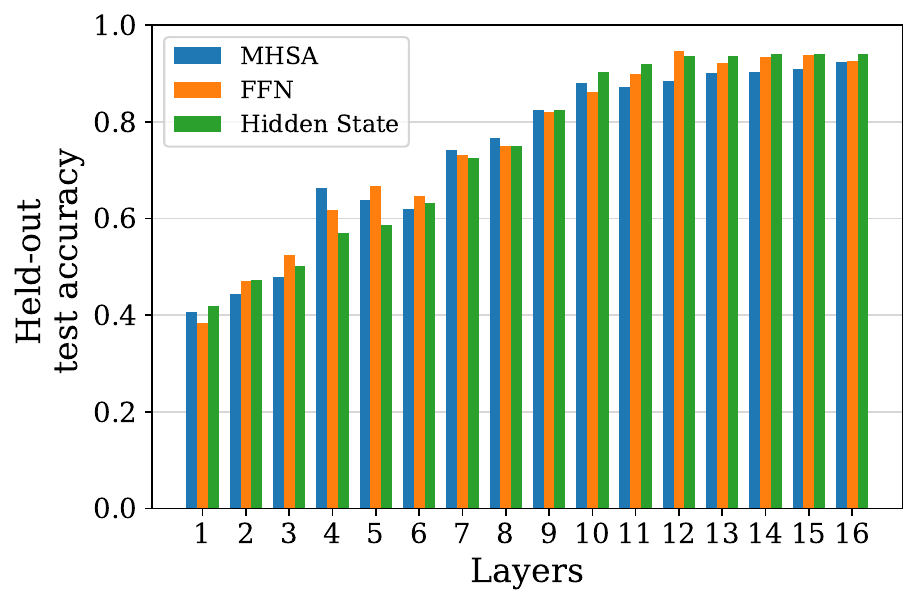}
    
  \caption{Probing test accuracy on last token of llama~3.2~1B for all layers. \textbf{Top} linear probe, \textbf{bottom} non-linear probe. There is a noticeable increase in probe performance in early layers when using a simple non-linear probe.}
  \label{fig:Llama1B_probe}
\end{figure}

\begin{figure}[bht]
    \centering
    \includegraphics[width=1.0\columnwidth]{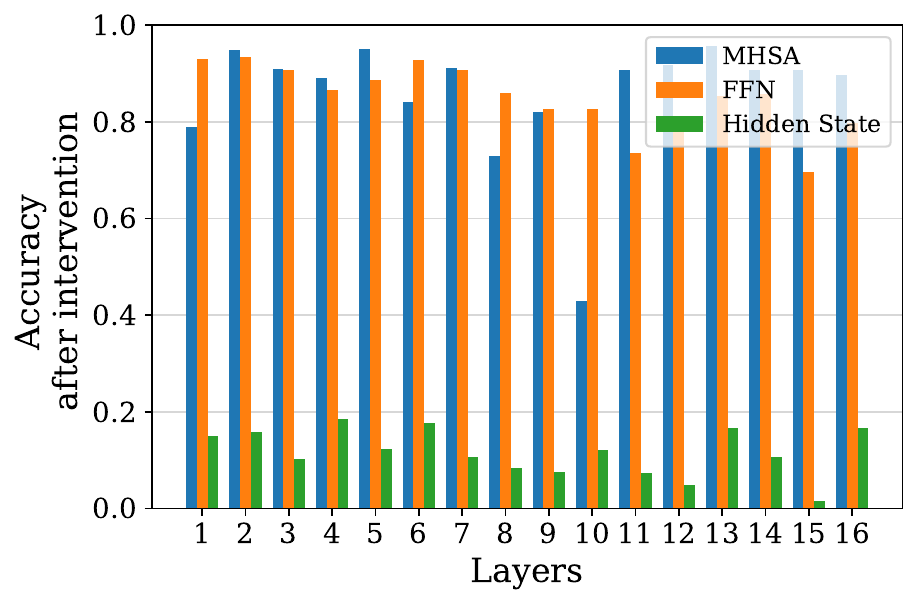}
  \caption{Zero-Intervention accuracy with span $1$ on the last token index of Llama~3.2~1B across all layers. There is a clear drop in accuracy when MHSA activations in layer 10 are knocked out.}
  \label{fig:Llama1B_zero}
\end{figure} 

\begin{figure}[bht]
    \centering
    \includegraphics[width=1.0\columnwidth]{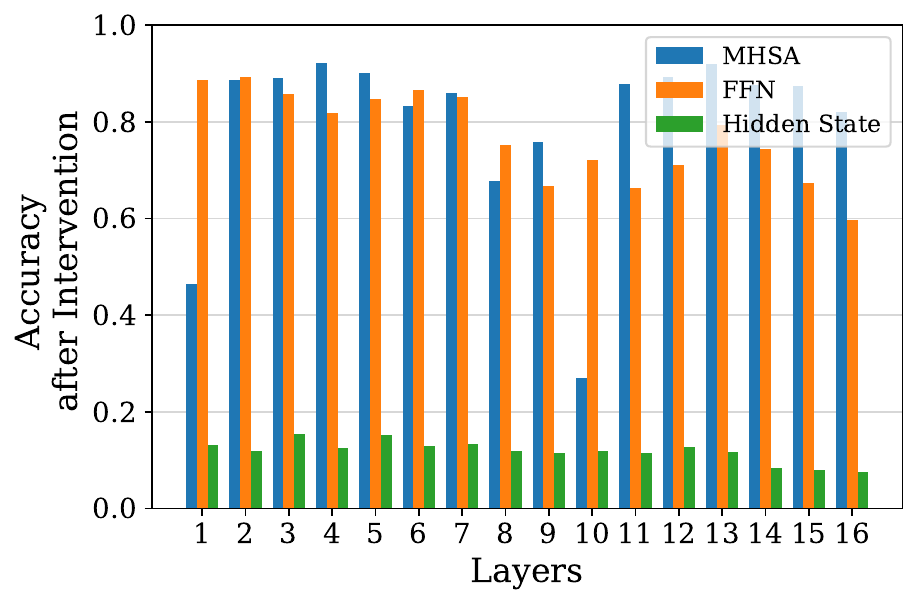}
  \caption{Random-Intervention accuracy with span $1$ on the last token index of Llama~3.2~1B across all layers. A noticeable drop in accuracy is observed when MHSA activations in layer 10 are knocked out.}
  \label{fig:Llama1B_random}
\end{figure}

\begin{figure*}[t!]
    \centering
    \includegraphics[width=1.0\linewidth]{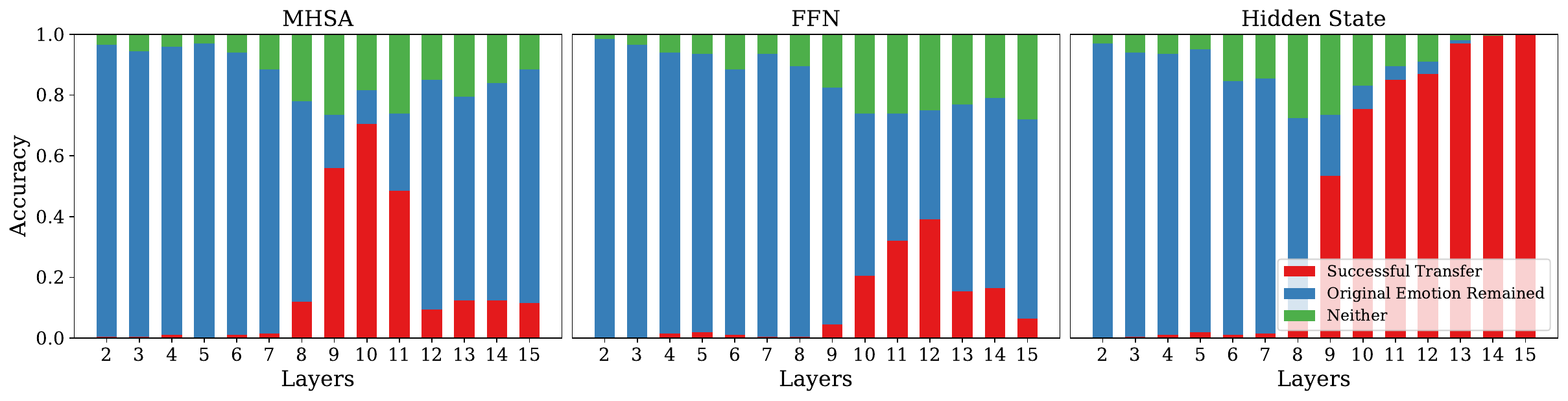}
    \caption{Activation patching results for Llama~3.2~1B across different layers (FFN, MHSA, and hidden state units) with Span = 3, evaluated over 200 source-target pairs. Blue indicates unsuccessful patching where the original label remained unchanged, red represents successful patching, and green denotes cases where the label changed but did not match the exact expected target.} 
    \label{fig:Llama1B_patching}
\end{figure*}

\begin{figure*}[t!]
    \centering
    \includegraphics[width=1.0\linewidth]{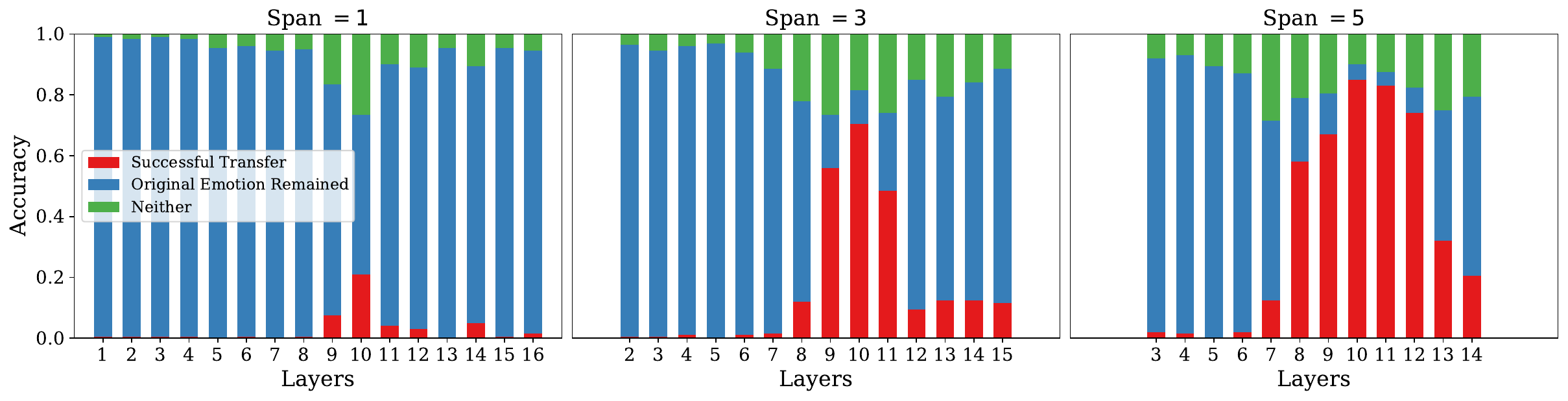}
    \caption{Effect of span size on Llama~3.2~1B activation patching at MHSA across different layers, evaluated over 200 source-target pairs.} 
    \label{fig:Llama1B_span_size}
\end{figure*}

\subsection{Details on Probing and Causal-Intervention} \label{app:detailed_figures}
Here, we report the results of probing and causal intervention experiments on the Llama 3.2 1B model, focusing on the last token index across all layers, along with a simplified illustration of the findings. Figure~\ref{fig:Llama1B_probe} top demonstrates linear probing results.

It is noteworthy that we use linear probes to detect and extract emotion vectors. Low probe accuracy on earlier layers does not mean that there is no emotion signal at early layers but rather suggests that the signal is not \textit{linearly} identifiable. In fact, Figure~\ref{fig:Llama1B_probe} bottom shows that when using a non-linear probe, i.e. a simple neural network with one hidden layer, the probe on earlier layers boosts considerably. However, both linear and non-linear probes peak around layer 10. After this layer, the model has finalized its output label decision and no major change happens. 

Intervention experiments further support this observation. Figures~\ref{fig:Llama1B_zero} and \ref{fig:Llama1B_random} show the effects of zero and random interventions with a span of 1, revealing a clear drop in accuracy when knocking out activations at layer 10. Finally, Figure~\ref{fig:Llama1B_patching} provides a detailed visualization of the patching experiment. The left-most plot highlights that the most successful emotion transfers occur at layer 10, which also exhibits the lowest number of unchanged labels. Notably, while some labels shifted to semantically similar emotions, they did not exactly match the target label and were, therefore, not counted as successful patches.

\subsection{Effect of Span Size on Activation Patching} \label{app:Span}
To examine the effect of span size on patching success, we repeated the experiment with three span sizes—1, 3, and 5 layers—on Llama 3.2 1B. Figure~\ref{fig:Llama1B_span_size} presents the results, showing a clear increase in patching effectiveness as the span size increases. Notably, layer 10 continues to exhibit peak patching performance, reinforcing the idea of emotion-related functional localization in that layer.

% Moreover, our findings suggest the existence of an optimal span size: small spans show negligible success, while excessively large spans risk disrupting unrelated functions in the model’s causal process. Identifying this balance is crucial for effective interventions, and future research should further investigate the trade-offs involved in span selection.

\subsection{Investigating Token Dimension} \label{app:token}

\begin{figure*}[bht]
    \centering
    \includegraphics[width=1.0\textwidth]{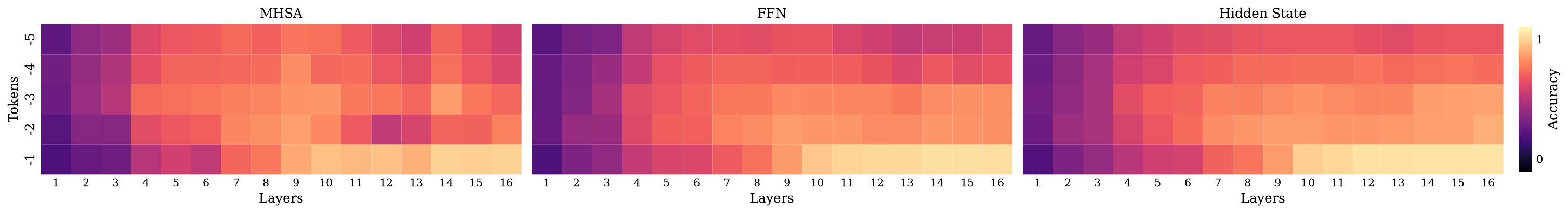}
  \caption{Probing test accuracy on different tokens of Llama~3.2~1B across all layers. We observe a consistent increase in signal strength from earlier to later tokens and from lower to higher layers.}
  \label{fig:Llama1B_probe_all_tokens}
\end{figure*} 

\begin{figure*}[bht]
    \centering
    \includegraphics[width=1.0\textwidth]{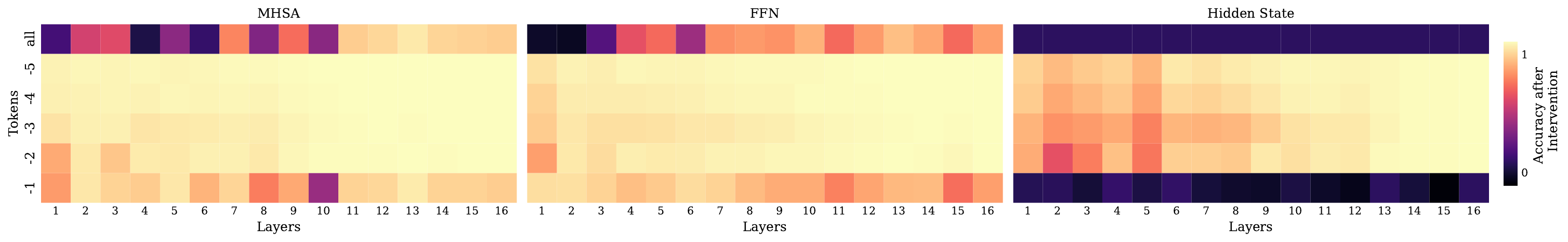}
  \caption{Zero intervention accuracy on different token indices of llama~3.2~1B for all layers with span = $1$. The vertical dimension shows tokens and \textit{"all"} means to knock out all activations of the specific layers at the same time. MHSA units beyond the critical layer (\( l = 10 \)) have minimal impact on the causal path, even when an entire layer is knocked out.}
  \label{fig:Llama1B_zero_all_tokens}
  % \vspace{-2mm}
\end{figure*} 

\begin{figure*}[bht]
    \centering
    \includegraphics[width=1.0\textwidth]{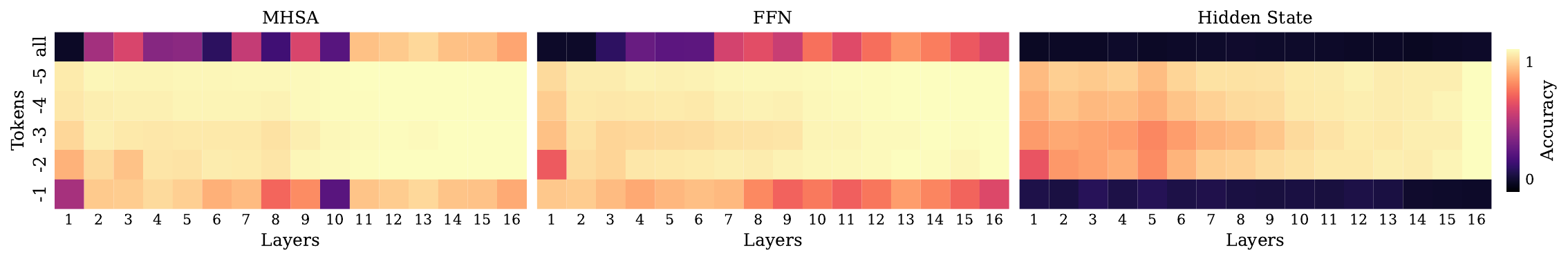}
  \caption{Random intervention accuracy on different token indices of Llama~3.2~1B across all layers with span = $1$. The vertical axis represents token positions, where \textit{"all"} denotes the simultaneous knockout of all activations in the specified layers. Notably, MHSA units beyond the critical layer (\( l = 10 \)) contribute minimally to the causal path, even when an entire layer is deactivated.}
  \label{fig:Llama1B_random_all_tokens}
\end{figure*} 

\begin{figure*}[bht]
    \centering
    \includegraphics[width=1.0\textwidth]{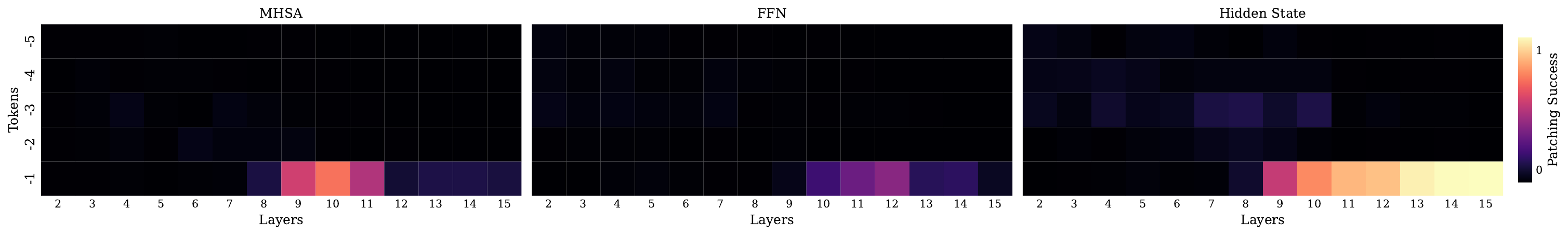}
  \caption{Results of activation patching success on different tokens of Llama~3.2~1B across all layers with span = 3. Clear functional localization is observed in layers 9–11.}
  \label{fig:Llama1B_patching_all_tokens}
\end{figure*} 

Throughout the paper, we extracted and analyzed the last token index activations $\mathbf{x}^{(l)}$---selected from one of the $\{ \mathbf{a}^{(l)}, \mathbf{m}^{(l)}, \mathbf{h}^{(l)} \}$---across different layers of all tested models. We hypothesized that the strongest emotion signals appear at the last token, as it directly influences the model's next-word prediction in a causal language modeling setup. Earlier work demonstrates the presence of strong causal states immediately before the prediction, as well as their emergence at the final token of a noun phrase \cite{meng2022locating}. Therefore, we suspect whether the last token of the query part in the prompt contains information more significant than the final token of the whole prompt.

To evaluate this hypothesis, we repeated the experiment on the last five tokens for Llama~3.2~1B extending the analysis to also include the final tokens in the query context (colored purple in Figure~\ref{fig:aggregate_attention_viz}-Top). In Figure~\ref{fig:Llama1B_probe_all_tokens}, we observe a consistent increase in signal strength from earlier to later tokens, reinforcing the focus on the last token as the primary contributor to output generation, but no clear importance on the last token of the query part. 

Similarly, we conduct zero-activation intervention, random-activation intervention, and activation patching on Llama~3.2~1B’s last five tokens across all layers. Figures~\ref{fig:Llama1B_zero_all_tokens}, \ref{fig:Llama1B_random_all_tokens}, and \ref{fig:Llama1B_patching_all_tokens} confirm our hypothesis, suggesting that the last token's MSHA units in mid-layers, particularly \({l = 10}\), are critical for processing emotional content, while other token positions exhibit no localization.

Furthermore, the first row of Figures~\ref{fig:Llama1B_zero_all_tokens} and \ref{fig:Llama1B_random_all_tokens} illustrates the effect of zero or random activation interventions applied to all token positions in a layer (i.e., when all units in a layer are knocked out). Notably, the results indicate that MSHA units beyond the critical layer \({l = 10}\) contribute minimally to the causal path, even when an entire layer is knocked out.

\subsection{Control Experiment on an Isomorphic Task} \label{app:control}

\begin{figure}[bht]
    \centering
    \includegraphics[width=1.0\columnwidth]{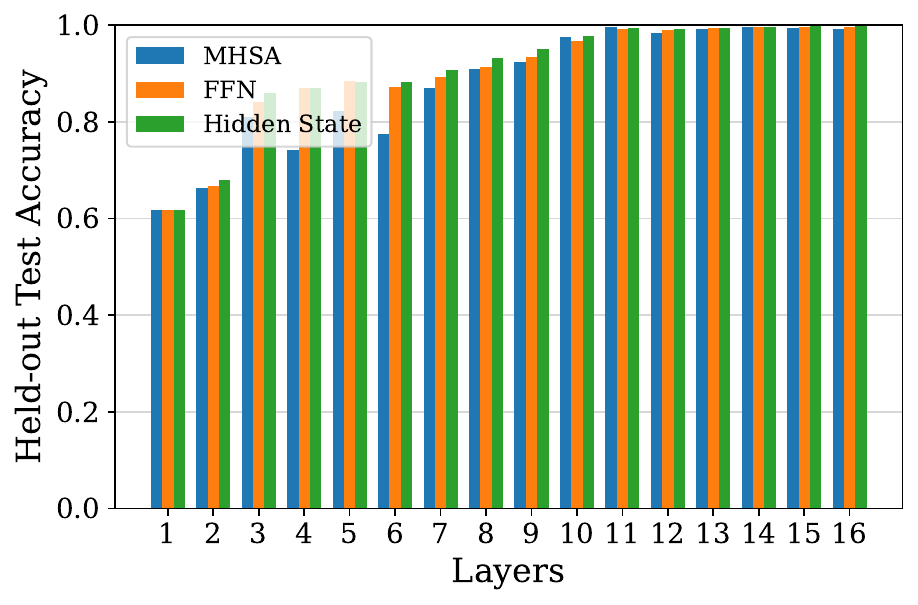}
  \caption{Probing test accuracy of the control isomorphic experiment on the last token of Llama~3.2~1B across all layers.}
  \label{fig:Llama1B_control_probe}
  % \vspace{-3mm}
\end{figure} 

\begin{figure*}[t!]
    \centering
    \includegraphics[width=1.0\linewidth]{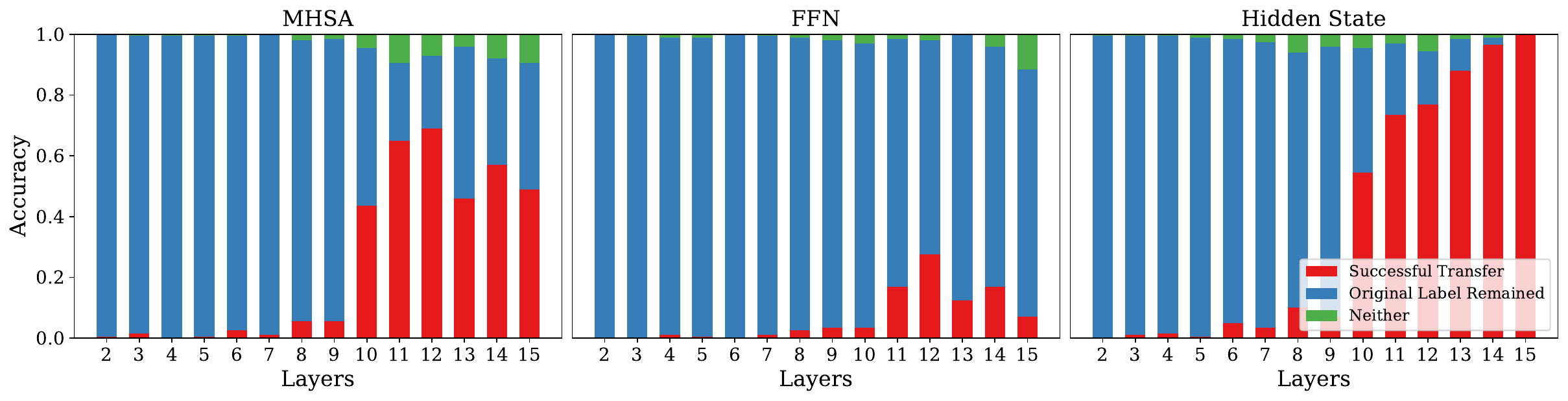}
    \caption{Control isomorphic experiment results for Llama~3.2~1B activation patching across different layers at MHSA, FFN, and hidden states with span = 3, evaluated over 200 source-target pairs. Localization is less evident, with the highest patching performance observed around the final layers.} 
    \label{fig:Llama1B_control_patching}
\end{figure*}

Here, we investigate whether the observed outcomes could be attributed solely to syntactic features and task structure rather than the target task of emotion processing. To assess this, we conduct an isomorphic experiment in which we modify the task to focus purely on syntax—predicting the first word in the sequence—and repeat the activation patching procedure as described in Section~\ref{sec:patching}. The altered prompt is shown below:

\begin{quote}
\textbf{What is the first word in the following contexts?}  
Context: My dog died last week. Answer: \textbf{My};   
Context: I saw moldy food. Answer: \textbf{I};   
Context: I could see my friend after a long time. Answer:
\end{quote}

As shown in Figures~\ref{fig:Llama1B_control_probe} and \ref{fig:Llama1B_control_patching}, MHSA units in the final layers of the model are most critical for this syntactic task, which contrasts significantly with the emotion patching findings. Additionally, we observe weaker evidence of functional localization based on the patching results from the isomorphic control experiment.

\subsection{Robustness to Prompt Design} \label{app:prompt}

\begin{figure*}[bht]
    \centering
    \includegraphics[width=0.7\textwidth]{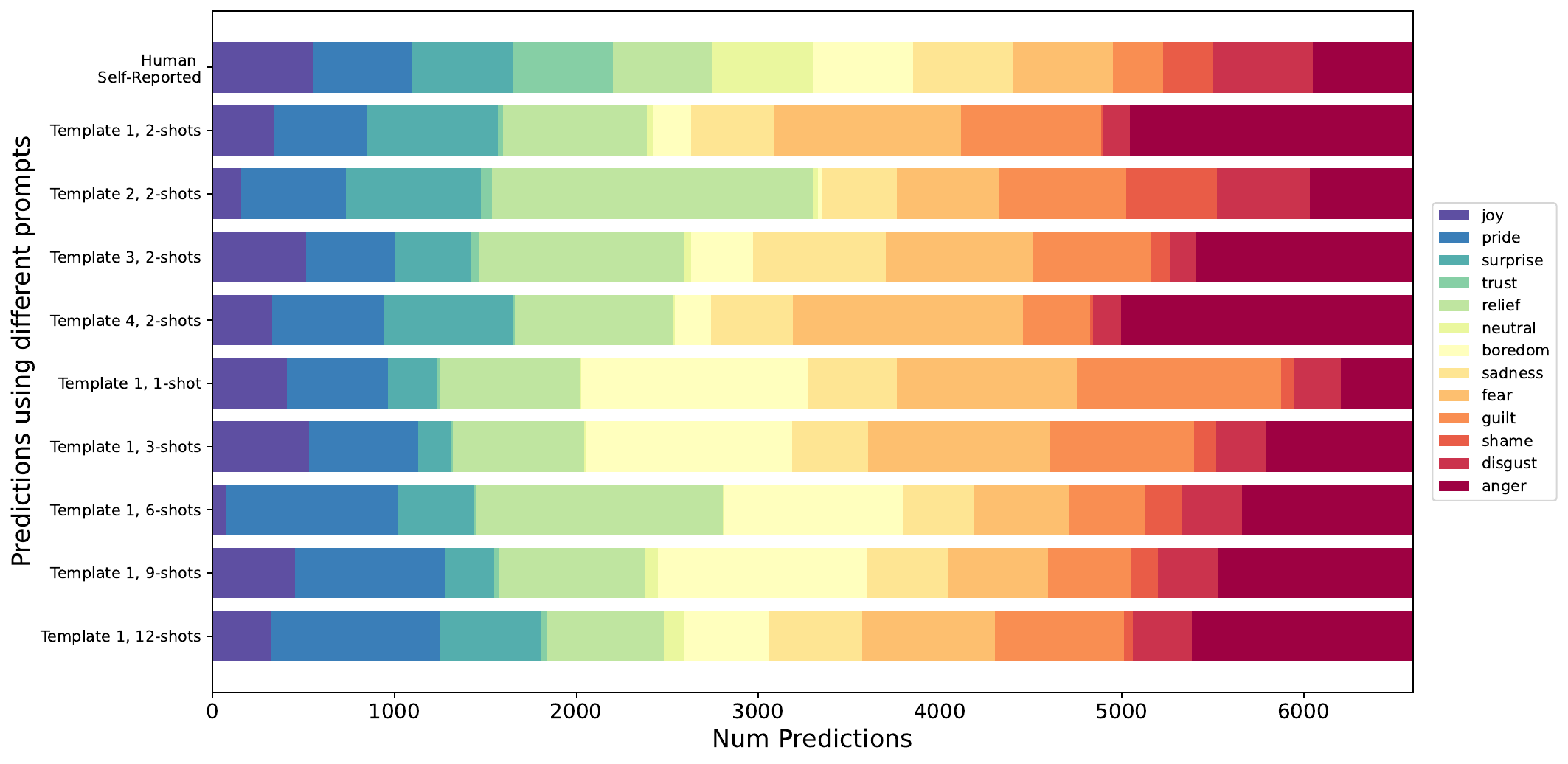}
  \caption{The distribution of next word emotion label predictions for different prompt templates and varied numbers of few shot examples.}
  \label{fig:prompt_distribution}
  % \vspace{-3mm}
\end{figure*}

\begin{figure}[bht]
    \centering
    \includegraphics[width=1.0\columnwidth]{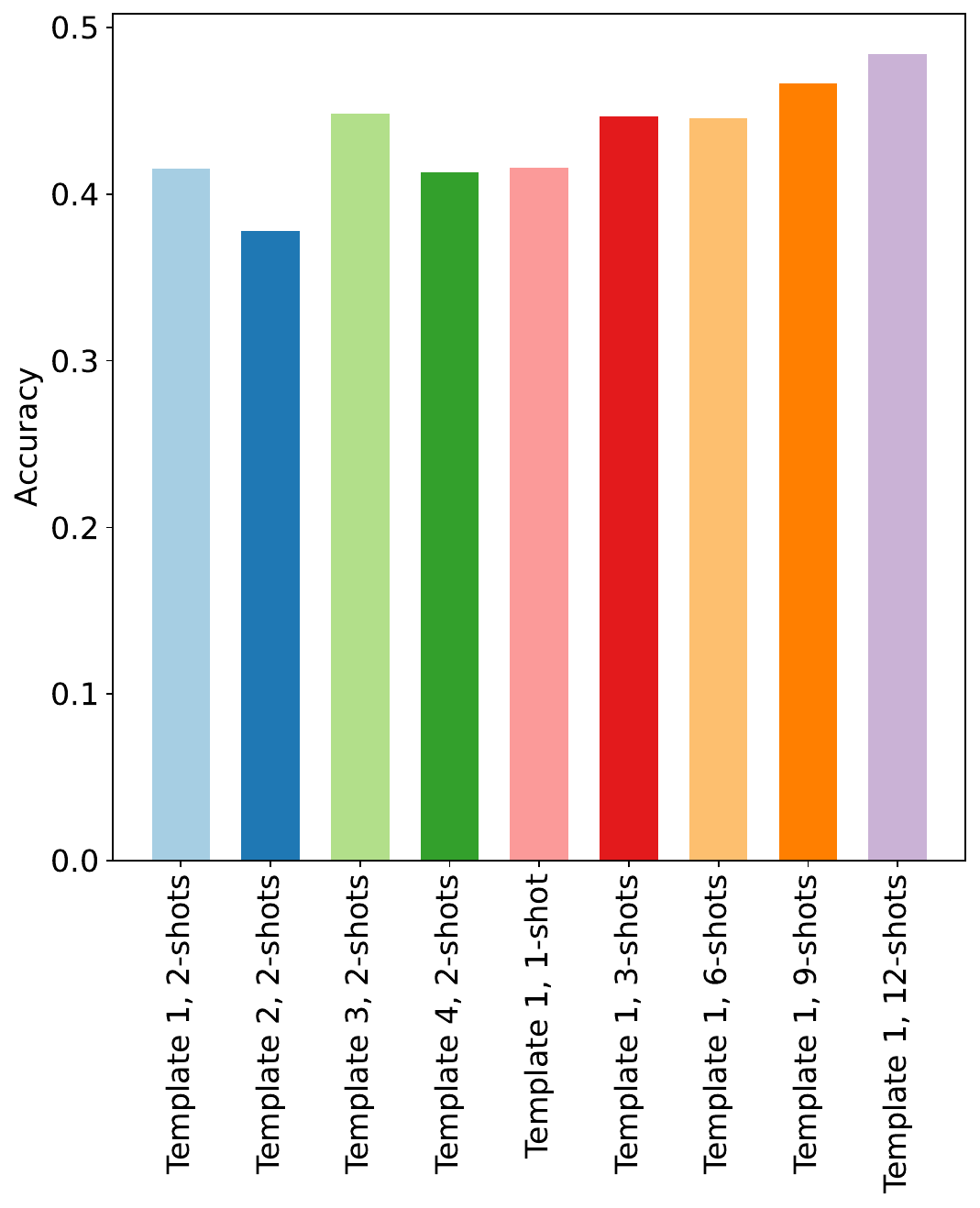}
  \caption{Accuracy of different prompts on the emotion classification task. This experiment varies both the prompt template and the number of provided few-shot examples. Experiments are conducted on Llama 3.2 1B.}
  \label{fig:prompt_accs}
  % \vspace{-3mm}
\end{figure}

To evaluate the model's robustness to various prompts, we designed a variety of prompt templates, as illustrated in Table \ref{table:prompt_templates}.
Figure \ref{fig:prompt_distribution} demonstrates the distribution of the next word emotion label predictions for the different prompt templates and different numbers of few-shot examples. Figure \ref{fig:prompt_accs} demonstrates the final accuracy of these tests. Noteworthy that again in the following interpretability experiments, we confine our focus only on the samples that the model could predict correctly using each specific prompt.

\begin{figure*}[bht]
    \centering
    \includegraphics[width=1.0\textwidth]{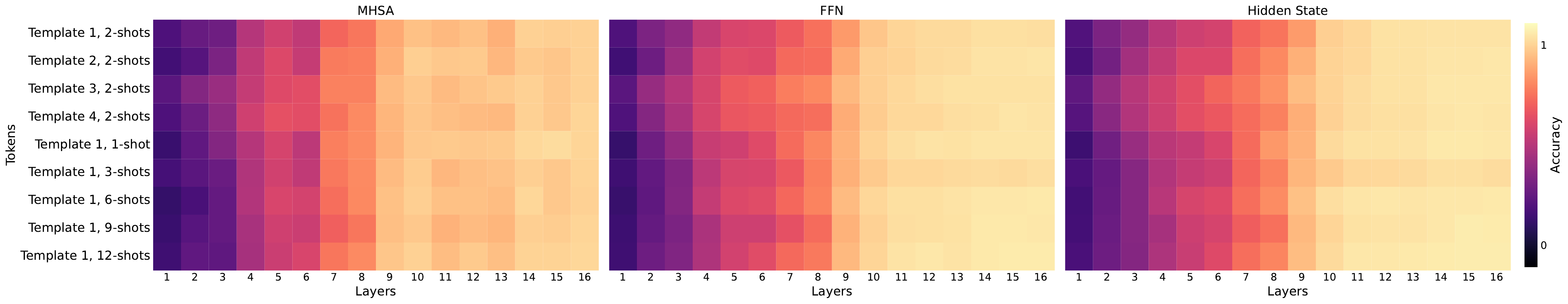}
  \caption{Probing accuracy with different prompts measured at last token in Llama~3.2~1B for all layers. This experiment varies both the prompt template and the number of provided few-shot examples.}
  \label{fig:prompt_probe}
\end{figure*}

\begin{figure*}[bht]
    \centering
    \includegraphics[width=1.0\textwidth]{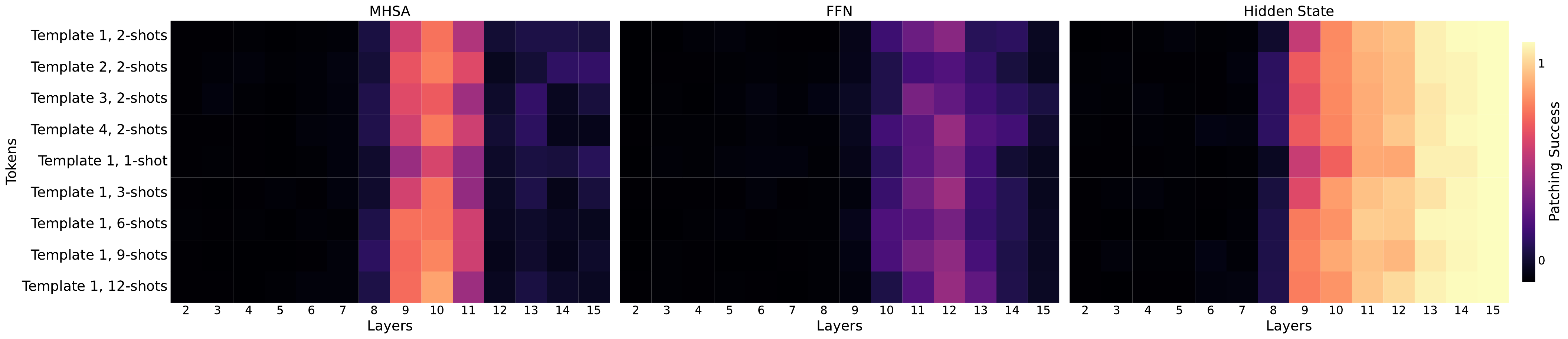}
  \caption{Success of activation patching with different prompts, measured at the last token index in Llama~3.2~1B across all layers with span = 3. This experiment varies both the prompt template and the number of provided few-shot examples.}
  \label{fig:prompt_patch}
\end{figure*}

Figures \ref{fig:prompt_probe} and \ref{fig:prompt_patch} validate that the probing and patching results presented earlier in the paper are robust to various prompt templates. For probing, we see that the model has determined the predicted label by the mid layers of the model, a result that is consistent across various prompt templates and the number of few shot examples provided. For activation patching, we also get consistent results across various prompt templates and with varied numbers of few-shot examples.

%mention figures \ref{fig:prompt_accs}, \ref{fig:prompt_distribution}, \ref{fig:prompt_probe}, \ref{fig:prompt_patch}.
%Also, please add these prompt templates into a table and mention we used them:

\begin{table}[ht]
% \centering
\tiny
\begin{tabular}{|p{0.02\columnwidth}|p{0.85\columnwidth}|}
\hline
\# & Prompt Template                                                                                                                                                                                                                                                                                                                                                              \\ \hline
1      & \begin{tabular}[c]{@{}l@{}}What are the inferred emotions in the following contexts?\\ Context: My first child was born.\\ Answer: joy\\ Context: My dog died last week.\\ Answer: sadness\\ Context: {[}Input{]}  \\ Answer:\end{tabular}                                                                                                                                       \\ \hline
2      & \begin{tabular}[c]{@{}l@{}}Consider this list of emotions: anger, boredom, disgust, fear, guilt, joy,\\ pride, relief, sadness, shame, surprise, trust, neutral.\\ What are the inferred emotions in the following contexts?\\ Context: My first child was born.\\ Answer: joy\\ Context: My dog died last week.\\ Answer: sadness\\ Context: {[}Input{]}  \\ Answer:\end{tabular} \\ \hline
3      & \begin{tabular}[c]{@{}l@{}}Context: My first child was born.\\ Answer: joy\\ Context: My dog died last week.\\ Answer: sadness\\ Context: {[}Input{]} \\ Answer:\end{tabular}                                                                                                                                                                                                   \\ \hline
4      & \begin{tabular}[c]{@{}l@{}}Guess the emotion.\\ Context: My first child was born.\\ Answer: joy\\ Context: My dog died last week.\\ Answer: sadness \\ Context: {[}Input{]}  \\ Answer:\end{tabular}                                                                                                                                                                             \\ \hline
\end{tabular}
\caption{The prompt templates used for experimenting with the language models. The [Input] would be replaced with the sample sentence from the dataset that we are trying to label.}
\label{table:prompt_templates}
% \vspace{-3mm}
\end{table}

% {\color{red}
% Template 1: What are the inferred emotions in the following contexts? Context: My first child was born. Answer: joy Context: My dog died last week. Answer: sadness Context: \{\_\_PLACEHOLDER\_\_\} Answer:

% Template 2: Consider this list of emotions: anger, boredom, disgust, fear, guilt, joy, pride, relief, sadness, shame, surprise, trust, neutral. What are the inferred emotions in the following contexts? Context: My first child was born. Answer: joy Context: My dog died last week. Answer: sadness Context: \{\_\_PLACEHOLDER\_\_\} Answer:

% Template 3:  Context: My first child was born. Answer: joy Context: My dog died last week. Answer: sadness Context: \{\_\_PLACEHOLDER\_\_\} Answer:

% Template 4: Guess the emotion. Context: My first child was born. Answer: joy Context: My dog died last week. Answer: sadness Context: \{\_\_PLACEHOLDER\_\_\} Answer:
% }

\subsection{Investigating the Incorrectly Classified Samples} \label{app:mistakes}

As mentioned earlier, we only used the dataset instances on which the LLM could predict the self-reported emotion label with no mistakes. We initially adopted this approach to reduce noise in the dataset, given the subjective nature of emotion labeling. Emotion classification can become a very subjective task in some contexts. For example, in many cases, it is hard to distinguish anger from guilt, given the short description provided by the participants. Please see Figure~\ref{fig:Llama1B_confusion_matrix}, where we show how closely related emotions might be confused. We emphasize that focusing only on correctly classified examples may introduce bias, particularly a bias toward unambiguous emotional content. This is common in much of the prior work in mechanistic interpretability, which focuses on tasks that LLMs perform well on to ensure reliable interpretation.

That being said, analyzing misclassified examples could offer a deeper understanding of model mechanisms. In this section, we conduct a preliminary analysis comparing appraisal patterns between correctly and incorrectly classified cases for each emotion. Interestingly, we found that LLMs tend to exhibit distinct appraisal profiles in misclassified cases, suggesting shifts in the underlying reasoning process.

Table~\ref{tab:misclas} summarizes these findings for three emotions that showed high confusion, presenting the mean and standard deviation of key appraisal dimensions using the appraisal probes we trained, along with t-test comparisons. These results suggest that misclassifications are not simply random errors but may stem from meaningful shifts in how the model evaluates appraisal cues.

\begin{table}[htbp]
\small
\centering
\begin{tabular}{lcc}
\hline
Emotion & \begin{tabular}[c]{@{}c@{}}Pleasantness \\ (Correct / Miss)\end{tabular} & \begin{tabular}[c]{@{}c@{}}Others agency \\ (Correct / Miss)\end{tabular} \\ \hline
\textit{Sadness} & $1.3 \pm 0.5 / 1.6 \pm 0.7$                                                      & $2.2 \pm 0.7 / 3.2 \pm 0.9$                                                       \\
\textit{Joy}     & $4.5 \pm 0.5 / 4.2 \pm 0.7$                                                      & $3.5 \pm 0.7 / 3.1 \pm 0.8$                                                       \\
\textit{Guilt}   & $1.9 \pm 0.6 / 1.6 \pm 0.8$                                                      & $2.8 \pm 0.7 / 3.3 \pm 0.7$                                                       \\ \hline
\end{tabular}
    \caption{Comparing the average appraisal values the LLM perceived for correctly classified vs misclassified samples with different emotion labels. All values reported here have p-value less than $0.001$ }
    \label{tab:misclas}
\end{table}

\section{Direct Emotion Promotion} \label{app:emotion_promotion}

In Section~\ref{sec:appraisal_intervention}, we showed that it is possible to change the model's output by manipulating appraisal concepts and directing it toward emotions with certain specifications of appraisals. In this section, we show that one can also directly inject a desired specific emotion label output by linearly adding the corresponding emotion vector to the hidden state of the model. More formally, recall the weight vector $\mathbf{w}_e$ for emotion $e$ as introduced in Section~\ref{sec:app_emo_projection}. We define the emotion promotion modification as 

\[
    \mathbf{x} \leftarrow \mathbf{x} + \beta \frac{\mathbf{w}_e}{\norm{\mathbf{w}_e}_2},
\]

where $\mathbf{x}$ on the RHS is the activation from the original model at any desired layer or location, and $\beta$ is the scaling factor that controls the strength of emotion promotion.

\begin{figure*}[htbp]
    \centering
    \includegraphics[width=1.0\textwidth]{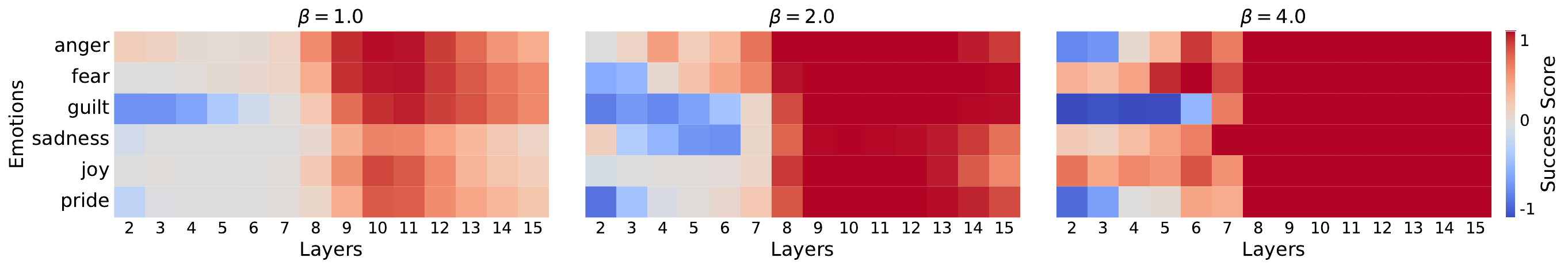}
  \caption{The heatmap showing the success of direct emotion promotion when applied at different layers of Llama~3.2~1B. Success score 1 means that the intervention successfully changed all output labels to the target emotion label. Scores 0 means that the intervention made no changes to the output and -1 means that the intervention resulted to complete opposite results, even damaging the samples with the correct original label. All the interventions in this figure used an intervention of layer span size 3.}
  \label{fig:emotion_promotion}
\end{figure*}

Figure~\ref{fig:emotion_promotion} shows the results of direct emotion promotion when performed on the hidden states across different layers of Llama~3.2~1B for different target emotion labels. As the figure suggests, direct emotion promotion is not effective when applied to the early layers, which completely aligns with our prior results. However, after layer 9, the success chance greatly improves, especially for large enough values of $\beta$. Again, this is a validation of our previous findings which shows that emotion concepts are linearly accessible and modifiable after the mid-layers. But before these layers, even a direct modification may fail since it will be overwritten later by the subsequent layers.

\section{Appraisal Probing} 
\label{app:appraisal_probing}

\begin{figure*}[t!]
    \centering
    \begin{subfigure}[t]{1.0\textwidth}
    \centering
    \includegraphics[width=1.0\linewidth]{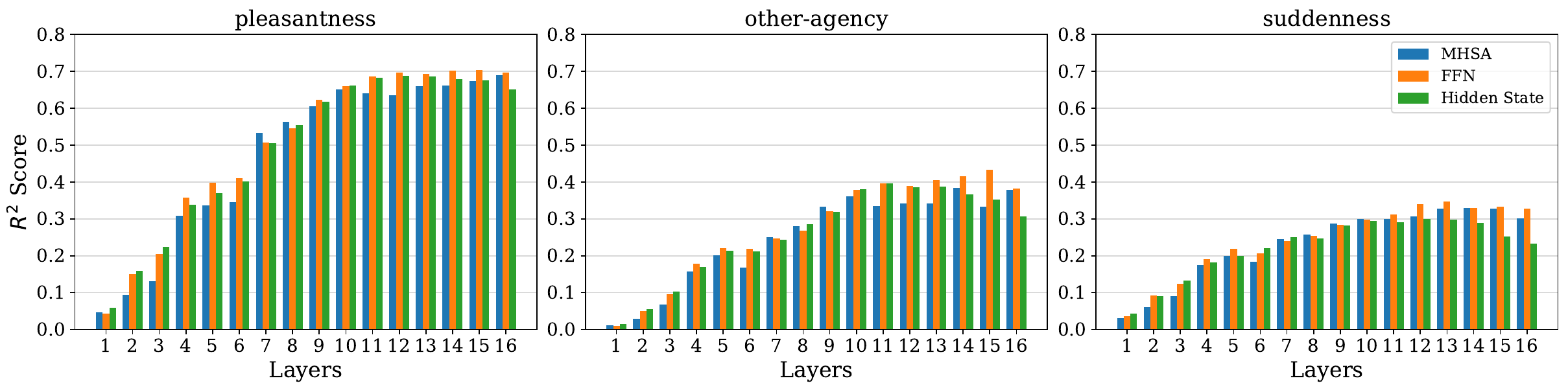}
    \end{subfigure}%
    \newline
    \newline
    \newline
    \begin{subfigure}[t]{1.0\textwidth}
        \centering
        \includegraphics[width=1.0\linewidth]{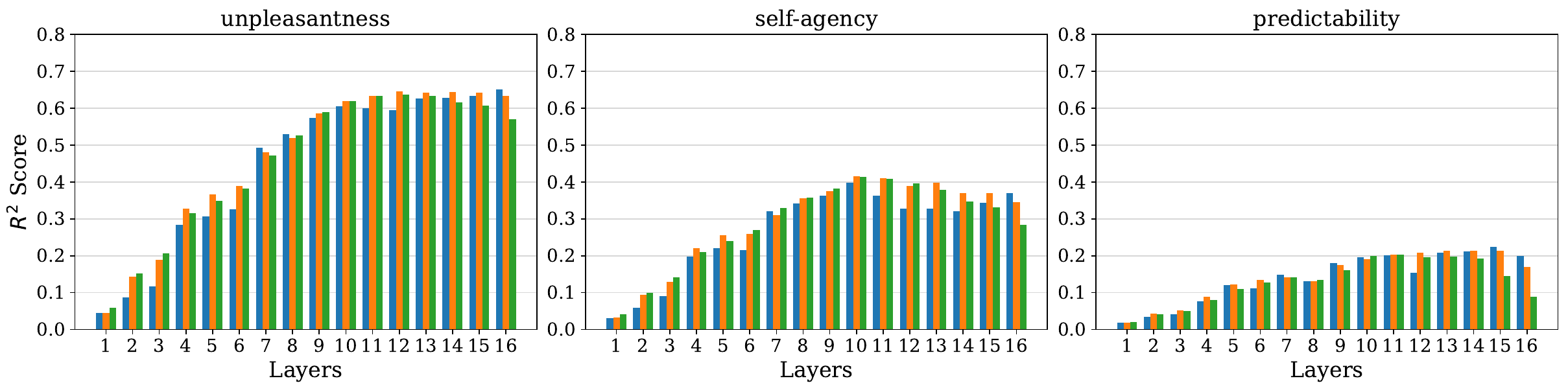}
    \end{subfigure}
    \caption{Probing results for Llama~3.2~1B, conducted separately for each appraisal dimension across different activation locations and layers for the last token. Results are measured as the regression $R^2$ score on a held-out test set.} 
    \label{fig:Llama1B_app_probing}
\end{figure*}

As detailed in Section~\ref{sec:appraisal_modulation} and similar to emotion probing experiments provided in Section~\ref{sec:emotion_probe}, we perform probing experiments to assess the presence and strength of appraisal signals at different activations within the Llama~3.2~1B model. We perform probing separately for each appraisal dimension over different activation locations and layers across the model for the last token. The probing results in Figure~\ref{fig:Llama1B_app_probing} are measured as the regression $R^2$ score on a held-out test set.

Following the behavior observed in probing emotion signals, models begin consolidating appraisal-related information in the hidden states $\mathbf{h}^{(l)}$ around the mid-layers. We observe that beyond layer $10$, there is no significant increase in probe accuracy in any appraisal dimension. As also observed in the emotion probing experiment, there is no clear distinction in probing performance between \( \mathbf{m}^{(l)} \) and \( \mathbf{h}^{(l)} \). 

As discussed in length in the main test, the success of linear probing highly depends on whether the target concept is linearly detectable given an activation. The results here also enforce the notion that the appraisal signals are not linearly detectable at earlier layers but are strongly present as we approach the hidden state of the final layers.

% \clearpage
\section{Further Details on Appraisal Modulation} \label{app:appraisal_modulation}

\begin{figure*}[bht]
    \centering
    \includegraphics[width=1.0\textwidth]{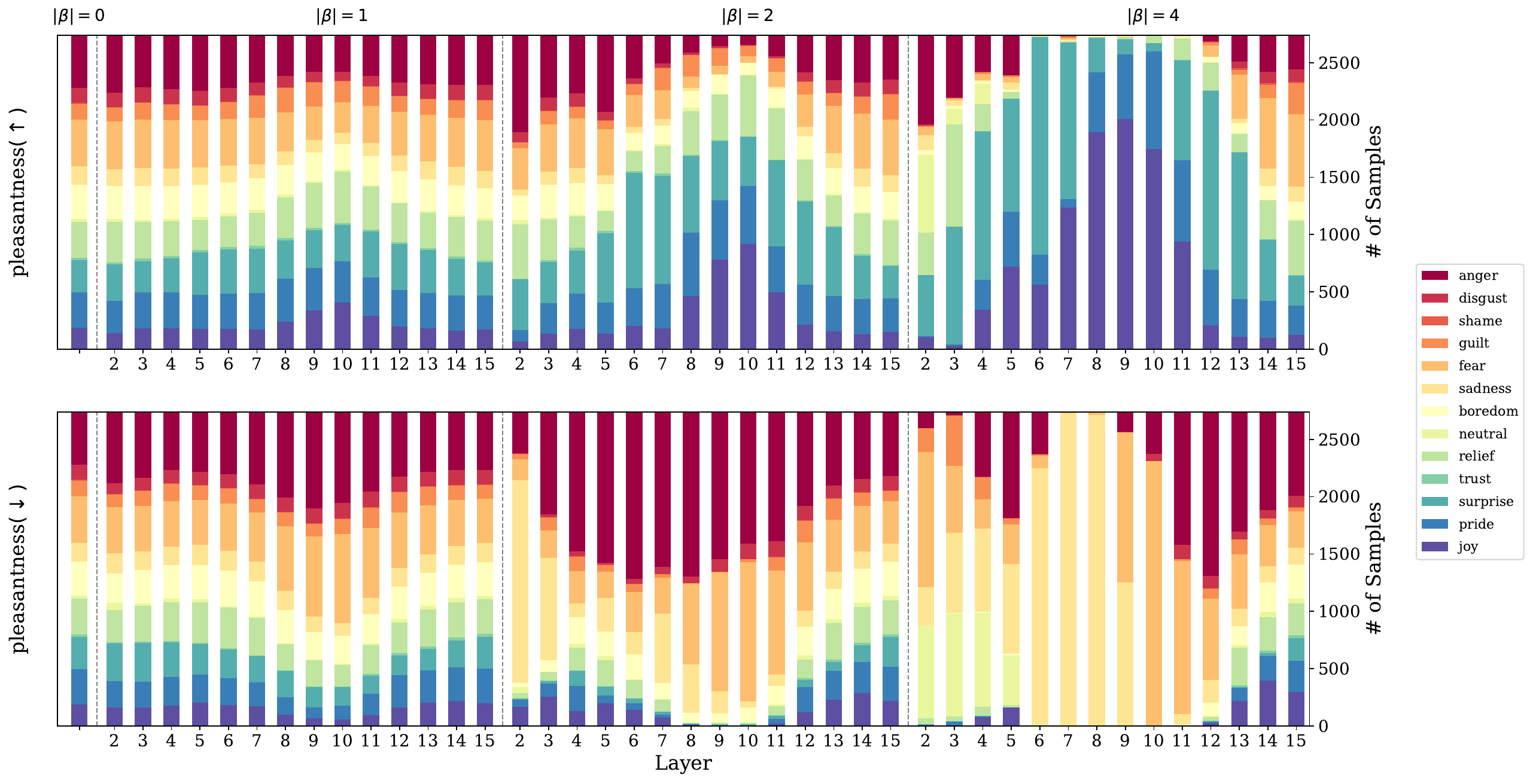}
  \caption{Effect of promoting and demoting \textit{pleasantness} at different layers of Llama~3.2~1B with three levels of scaling factor $\beta$. $\beta = 0$ represents the original distribution without appraisal modulation. A consistent increase in distribution shift is observed as $\beta$ increases across all intervention experiments. However, when intervening on earlier layers, particularly at higher $\beta$ values, the shift in the distribution of represented emotions does not always align with theoretical expectations.}
  \label{fig:appraisal_layers_pleasantness}
\end{figure*} 

\begin{figure*}[htbp]
    \centering
    \includegraphics[width=1.0\textwidth]{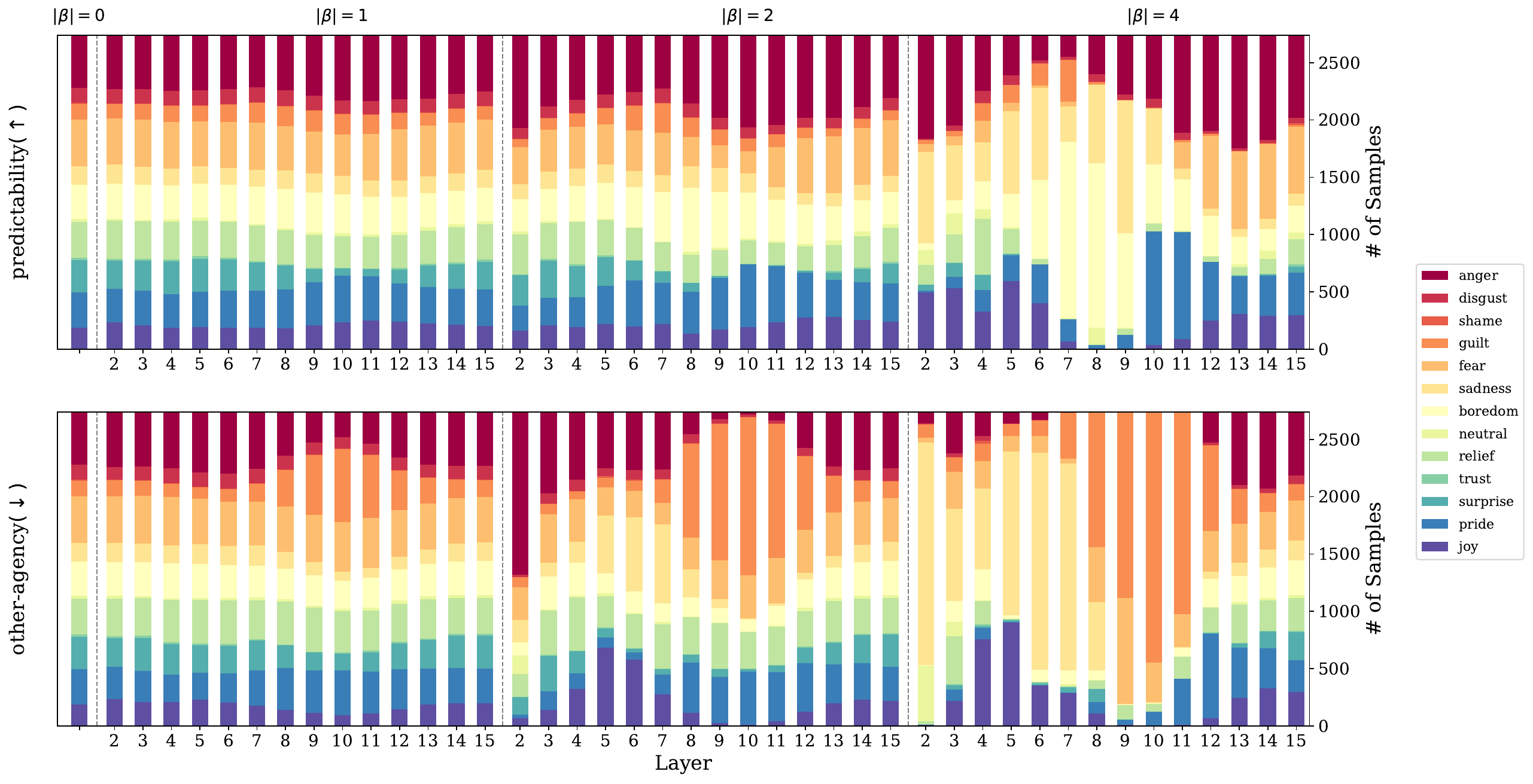}
  \caption{Effect of promoting and demoting \textit{other-agency} at different layers of Llama~3.2~1B using three levels of scaling factor $\beta$. $\beta = 0$ represents the original distribution without appraisal modulation. Mid-layer appraisal modulation exhibits a theoretically plausible shift in emotion distribution.}
  \label{fig:appraisal_layers_otheragency}
\end{figure*} 

\begin{figure*}[htbp]
    \centering
    \includegraphics[width=1.0\textwidth]{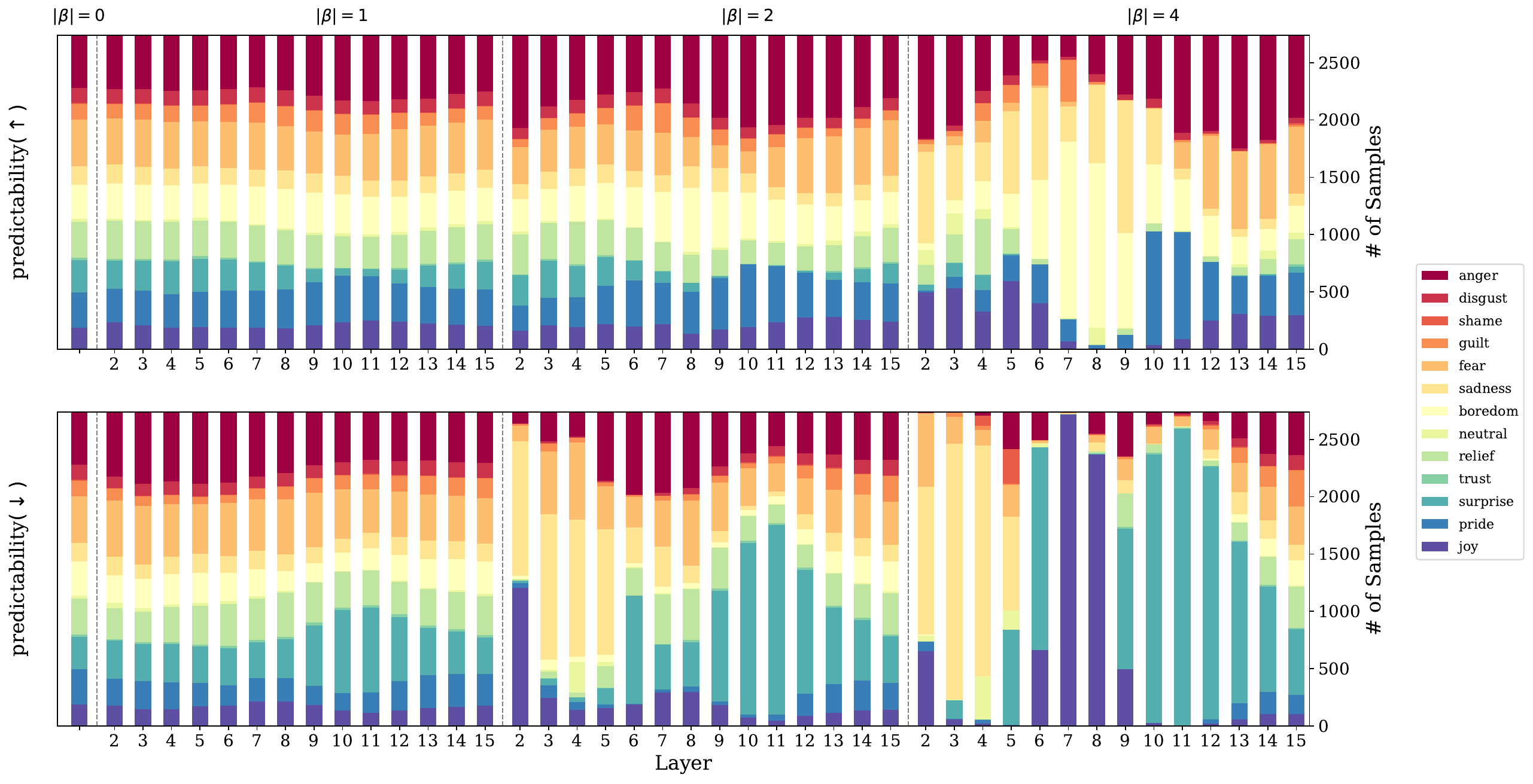}
  \caption{Effect of promoting and demoting \textit{predictability} at different layers of Llama~3.2~1B using three levels of scaling factor $\beta$. $\beta = 0$ represents the original distribution without appraisal modulation. Mid-layer appraisal modulation exhibits a theoretically plausible shift in emotion distribution.}
  \label{fig:appraisal_layers_predictability}
\end{figure*}

As discussed in Section~\ref{sec:appraisal_intervention}, we show the possibility of indirectly modifying the emotion of an input example by modulating its appraisals within the model representations. In this section, we provide more details and further experiments on appraisal modulation.

First, we redefine our appraisal modulation method to generalize to cases where we want to modify multiple concepts simultaneously. Let's assume that we have a set of $r$ appraisal vectors $\mathcal{A} := \{\mathbf{v}_{i_1}, \mathbf{v}_{i_2}, \cdots, \mathbf{v}_{i_r}\}$ which we want to modify, and consider a second set $\mathcal{B} := \{\mathbf{v}_{j_1}, \mathbf{v}_{j_2}, \cdots, \mathbf{v}_{j_k} \}$ to contain the $k$ appraisal vectors which we want to maintain the corresponding appraisal concept fixed during the update. We define the project matrix $\mathbf{P}_{\mathcal{B}}$ as the projection matrix which projects into the span of $\mathcal{B}$. More formally, form the matrix $\mathbf{V}_{\mathcal{B}} = [\mathbf{v}_{j_1}; \mathbf{v}_{j_2}; \cdots; \mathbf{v}_{j_k}] \in \mathbb{R}^{d \times k}$ by concatenating the vectors in $\mathcal{B}$ as the columns of $\mathbf{V}_{\mathcal{B}}$. With this in mind, the projection matrix, $P_{\mathcal{B}}$ as

\[
P_{\mathcal{B}} = \mathbf{V}_{\mathcal{B}}(\mathbf{V}_{\mathcal{B}}^\top \mathbf{V}_{\mathcal{B}})^{-1} \mathbf{V}_{\mathcal{B}}^\top,
\]

and define the net effect vector as 

\[
    \mathbf{z}_{\mathcal{A}} := (I - \mathbf{P}_{\mathcal{B}})\sum_{a=i_1}^{i_r} \gamma_a \mathbf{v}_a,
\]

where each $\gamma_a$ is a variable from $\{-1, +1\}$ to indicate if the modulation promotes concept $a$ or demote it. Finally, the modulation is performed as

$$\mathbf{x} \leftarrow \mathbf{x} + \beta \frac{\mathbf{z}_{\mathcal{A}}}{\norm{\mathbf{z}_{\mathcal{A}}}_2}$$

where $\beta \in \mathbb{R}_+$ is a positive scaling factor.After performing the intervention, we measure the intervention's success by evaluating the new emotion label obtained by this modification across all examples in the dataset.

Section~\ref{sec:appraisal_intervention} reported the inference-time intervention results targeting the hidden state at layer 9 in Llama~3.2~1B, as we showed it is a critical point in emotion processing in Llama~3.2~1B. Here, we report the intervention results across all layers for a varied set of appraisal dimensions and their superposition to create more complex but specific concepts. 

As shown in Figures~\ref{fig:appraisal_layers_pleasantness}, \ref{fig:appraisal_layers_otheragency}, and \ref{fig:appraisal_layers_predictability}, the intervention on appraisal concepts changes the distribution of the output labels and this distribution shift is intensified with higher values of
\( \beta \) in all intervention experiments. However, the shift in the distribution of represented emotions does not necessarily conform with theoretical and intuitive expectations when intervening on earlier layers, particularly noticeable when \( \beta \) is sufficiently large. For example, in Figure~\ref{fig:appraisal_layers_pleasantness}, we note an unexpected decrease in the distribution of \textit{joy} and \textit{pride} in early layers, whereas psychologically plausible manipulations—such as an increase in high-valence emotions like \textit{joy}, \textit{pride}, and \textit{surprise}, only emerge in mid-layers, peaking at layer 9. This observation supports the notion that intervening on the first layers is not effective because the linear structure in representations is not formed well yet. 

Observing the intervention effect on later layers, we see significantly less pronounced distribution shifts. This also supports our earlier finding that the intervention on final layers is not effective because of the orthogonality of concepts that we showed in Section \ref{sec:app_emo_projection}. 

\begin{figure*}[htbp]
    \centering
    \includegraphics[width=1.0\textwidth]{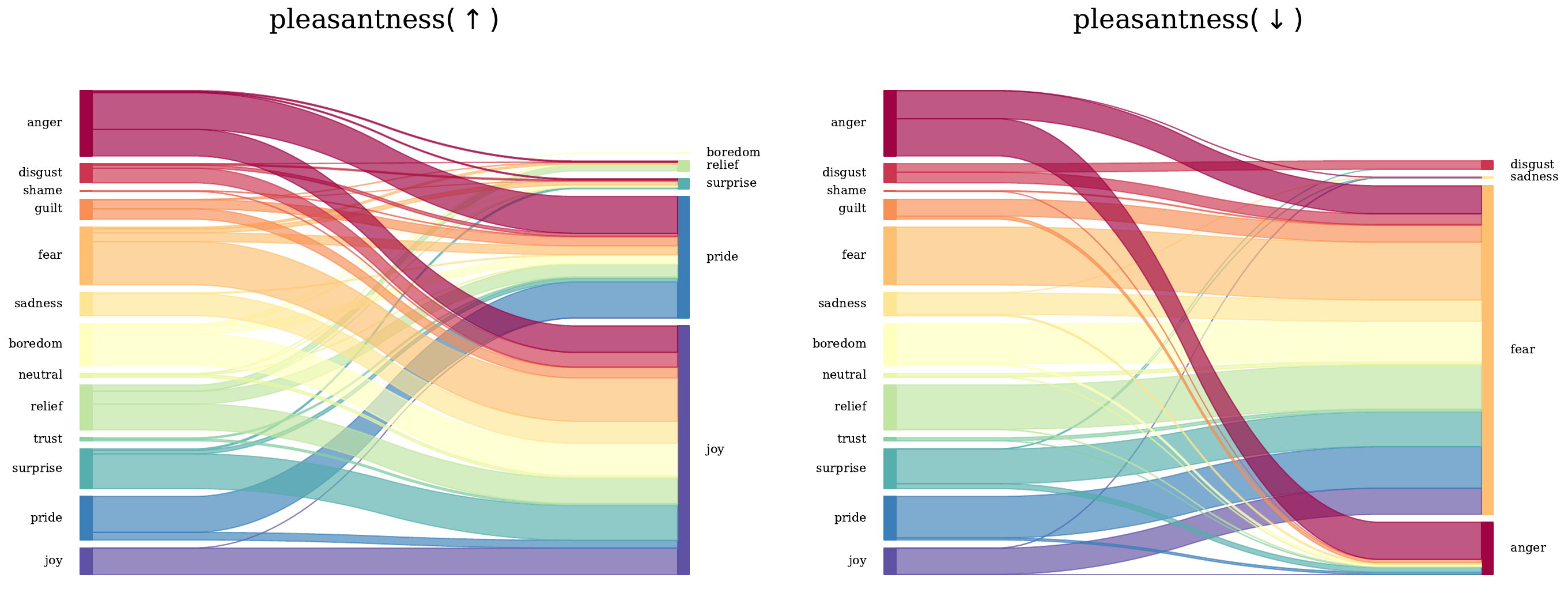}
  \caption{Sankey plot for \textit{pleasantness} appraisal modulation when we perform it at layer 9 of Llama~3.2~1B model.}
  \label{fig:appraisal_sankey_pleasantness}
\end{figure*}

On the other hand, We observe a remarkable alignment with theoretical and intuitive expectations in the distribution shifts associated with interventions on middle layers (specifically layers 9-11). For instance, we observe that increasing the \textit{pleasantness} appraisal promotes both \textit{joy} and \textit{pride}, aligning with the fact that both of these emotions have high associations with the appraisal. Also see Figure \ref{fig:appraisal_sankey_pleasantness} for a different visualization.

\begin{figure*}[htbp]
    \centering
    \includegraphics[width=1.0\textwidth]{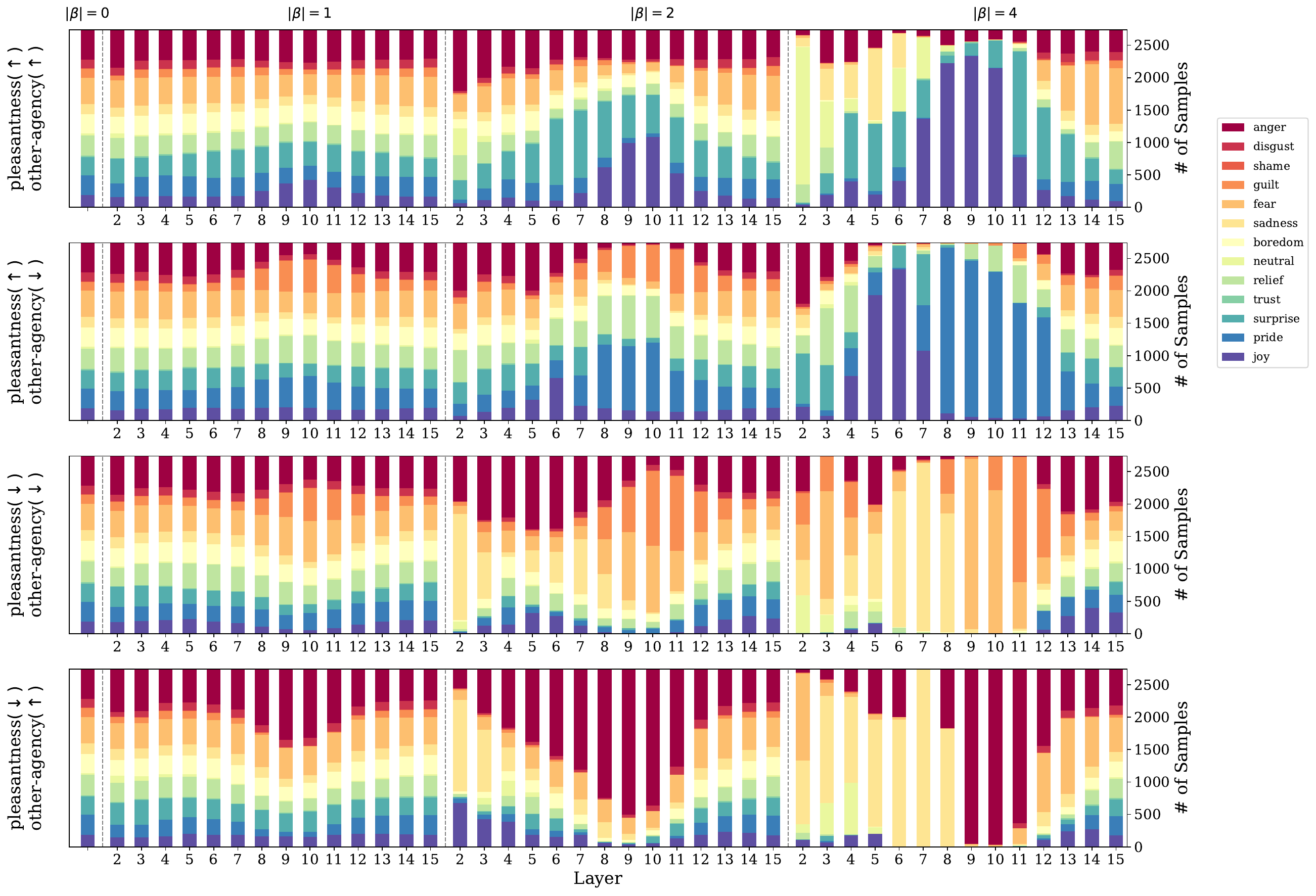}
  \caption{Superposition of \textit{pleasantness} and \textit{other-agency} appraisal modulation at different layers of Llama~3.2~1B. Results show successful promotion of \textit{pride} with no further occurrences of \textit{joy} in layers 9–11 when demoting \textit{other-agency}, and the promotion of \textit{guilt} and \textit{fear} with no occurrences of \textit{anger} in mid-layers when demoting \textit{other-agency}.}
  \label{fig:appraisal_layers_pleasantness_otheragency}
\end{figure*}

\begin{figure*}[htbp]
    \centering
    \includegraphics[width=1.0\textwidth]{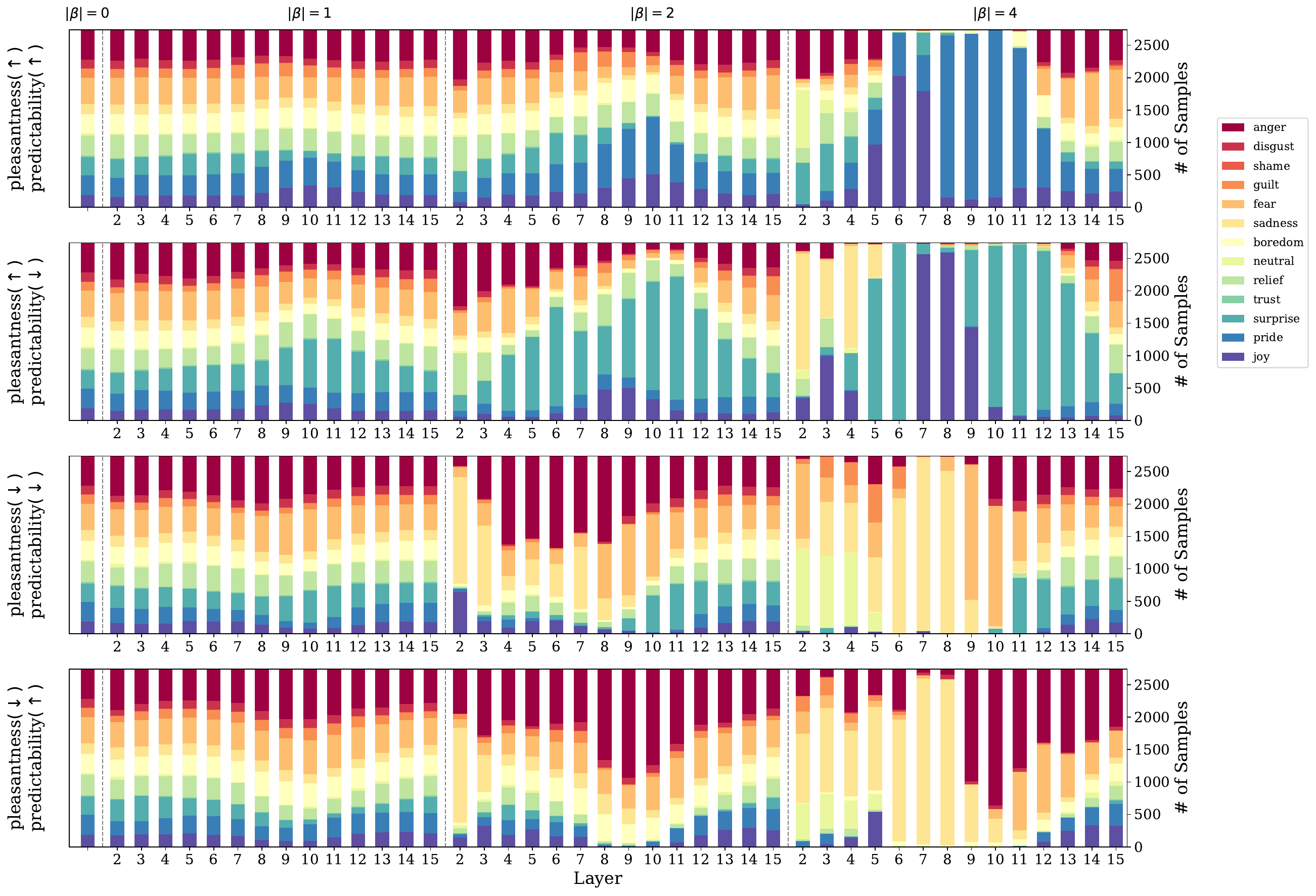}
  \caption{Superposition of \textit{pleasantness} and \textit{predictability} appraisal modulation at different layers of Llama~3.2~1B. Results show a successful transition from \textit{pride} to surprise and a greater distribution of \textit{fear} compared to anger in mid-layers when promoting unpredictability.}
  \label{fig:appraisal_layers_pleasantness_predictability}
\end{figure*}

To better evaluate intervention success, we also provide intervention results on the superposition of two appraisal dimensions (e.g., \textit{other-agency} and \textit{pleasantness}) across all layers in Llama~3.2~1B. The results, demonstrated in Figure~\ref{fig:appraisal_layers_pleasantness_otheragency}, show a successful promotion of emotion \textit{pride} with no further occurrences of \textit{joy} in layers 9-11 when demoting \textit{other-agency} and the promotion of \textit{guilt} and \textit{fear} with no occurrence of anger in mid-layers when demoting \textit{other-agency}. Similarly, in Figure~\ref{fig:appraisal_layers_pleasantness_predictability}, we see the transition from \textit{pride} to \textit{surprise} and a larger distribution of \textit{fear} as compared with \textit{anger} when we promote \textit{unpredictability}.

\begin{figure*}[htbp]
    \centering
    \includegraphics[width=1.0\textwidth]{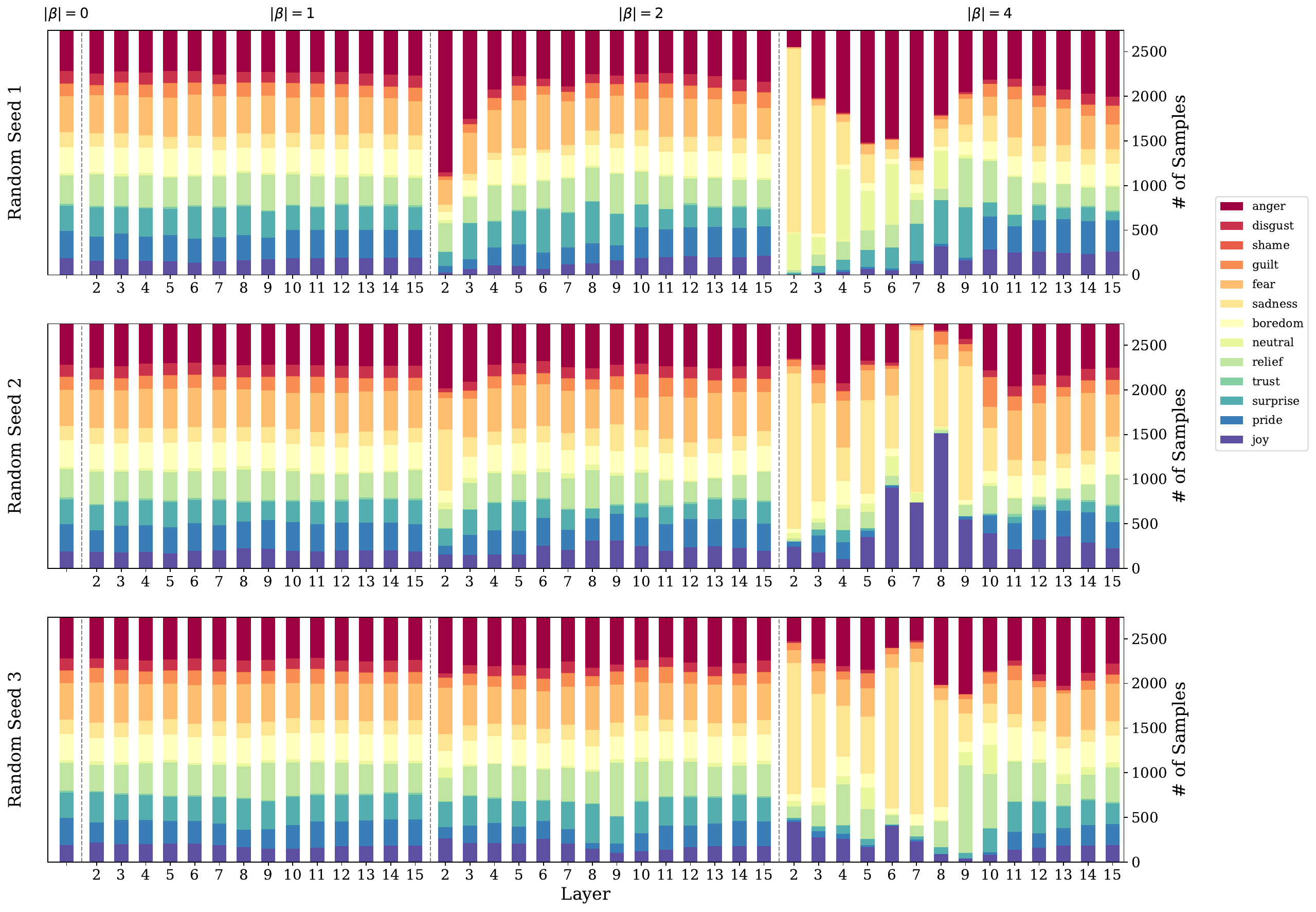}
  \caption{Results of the control experiment in which we randomly sample a vector and add it to the hidden state of the Llama~3.2~1B model at different layers. Each Row shows a different random seed.}
  \label{fig:appraisal_random}
\end{figure*}

Finally, we will conclude this section by providing a control experiment in which, instead of using appraisal vectors, we intervene in the activation using a random vector. The results of this experiment are provided in Figure~\ref{fig:appraisal_random}, and we see that no psychologically valid pattern is observable. Overall, these findings provide strong evidence that the mid-layers in Llama~3.2~1B meaningfully and directly contribute to cognitive processes related to emotions. 

% \clearpage
\section{Code and Compute Resources}
Our experiments are conducted using GPU-accelerated compute resources, with hardware such as NVIDIA A100 GPUs. For larger models, our studies are feasible on GPUs with at least 40GB of VRAM, with the full experiment running in approximately 24 hours. For smaller models, GPUs with 12GB of VRAM are sufficient to carry out our analyses efficiently. 
%Our implementation and experiment code are publicly available at:~\href{https://github.com/aminbana/emo-llm.git}{Emo-LLM Github Repository}.
% The code and data used in this study are attached as a supplementary file.

Generative AI tools are utilized to improve the tone and style of writing, as well as for code completion during implementation.

\end{document}